\documentclass[11pt, a4paper, arxiv]{google}

\usepackage[authoryear, sort&compress, round]{natbib}
\usepackage{graphicx}

\usepackage{bm}
\usepackage{accents}

\DeclareMathOperator*{\argmin}{arg\,min}

\newcommand{\KL}{\textrm{KL}}
\newcommand{\E}{\mathbb{E}}
\newcommand{\N}{\mathcal{N}}


\title{Spectral Souping: A Unified Framework for Online  Preference Alignment}

\correspondingauthor{yinlamchow@google.com}

\reportnumber{0001} 

\renewcommand{\today}{2026-05-19}

 \newtheorem{theorem}{Theorem}
 \newtheorem{assumption}{Assumption}
 \newtheorem{lemma}{Lemma}
 \newtheorem{remark}{Remark}

\author[1]{Yinlam Chow}
\author[2]{Guy Tennenholtz}
\author[1]{Ted Yun}
\author[1]{James Harrison}
\author[1]{Arthur Gretton}
\author[1]{Andre Barreto}
\author[1]{Bo Dai}

\affil[1]{Google DeepMind}
\affil[2]{Google Research}

\begin{abstract}
Reinforcement Learning from Human Feedback (RLHF) effectively aligns Large Language Models (LLMs) with aggregate human preferences but often fails to address the diverse and conflicting needs of individual users. To overcome this issue, we introduce \emph{Spectral Souping}, a unified framework for efficient, online  preference alignment. Our core contribution is the discovery of a universal spectral representation within LLMs, which is proven to be highly amenable to model merging. 
This theoretical insight enables a two-phase methodology: we first learn a basis of specialized policies offline, each focused on a distinct, fine-grained preference dimension. An online adaptation algorithm then efficiently ``soups'' these policies at inference time, either by merging their outputs or parameters, enabling rapid model adaptation without the need for costly online retraining w.r.t. tailored preference rewards. Experiments on online preference alignment benchmarks demonstrate that our method achieves significant performance improvements over existing state-of-the-art approaches, presenting a scalable and computationally efficient solution for dynamically adapting LLMs to individual user preferences.
\end{abstract}

\begin{document}

\maketitle

\section{Introduction}
Recent advancements in LLMs have demonstrated remarkable success in aligning with human preferences through techniques like RLHF \citep{ouyang2022training} and Direct Preference Optimization (DPO) \citep{rafailov2023direct}. However, these methods, which rely on a unified reward from aggregated feedback, face significant limitations. The core issue is that a one-size-fits-all approach fails to account for the diverse and often conflicting needs of individual users, which stem from differences in backgrounds and contexts. This gap between generalized and specialized preferences highlights a critical challenge of aligning LLMs with individual preferences without incurring the substantial data collection and computational costs associated with fine-tuning a separate model for each user.

Our work introduces \emph{Spectral Souping}, a novel framework for online personalized preference alignment of LLMs that overcomes these limitations. Unlike traditional methods, which require separate and costly fine-tuning, our method can handle the diverse and varying user preferences in an efficient way. Our core contribution lies in the discovery of a (universal) spectral representation in the context of the language Markov Decision Process (MDP), where the LLM policy maximizes the user's preference-driven reward. This observation shows that the logits of various personalized LLM policies do not exist in an arbitrary space but rather in a structured latent space defined by the MDP's spectral features, implying these logits can be represented as a linear combination of a small number of basis logit functions—each corresponds to a policy that aligns with a distinct preference dimension. 

This theoretical insight underpins our two-phase LLM-adaptation methodology. An offline phase trains the aforementioned basis of specialized policies. The online phase then dynamically combines these basis policies at inference time to generate responses tailored to user preferences, thereby obviating the need for costly per-user fine-tuning. The resulting framework is highly scalable and can achieve state-of-the-art performance on online preference alignment benchmarks. Critically, our discovery of a unified spectral representation enables the derivation of provable sub-optimality bounds for this policy merging approach—a significant advancement over prior techniques, which were largely heuristic and lacked such formal guarantees. In particular, we show that our spectral souping method achieves performance arbitrarily close to that of a fully fine-tuned ``tailored'' policy, thus complementing its empirical efficacy with a rigorous theoretical foundation.

The rest of this paper is organized as follows. Section \ref{sec:prelim} provides background on the language MDP, RLHF, and the online preference alignment problem. Section \ref{sec:universal_rep} details our theoretical discovery of the spectral representation and its properties related to LLM preference alignment. Section \ref{sec:alg} presents our two-phase methodology, including the offline training of specialized policies and the online souping algorithm. Section \ref{sec:exp} describes our experimental setup and presents our results, demonstrating the efficacy of spectral souping. Finally, Section \ref{sec:related_work} delineates related work on LLM adaptation, and Section \ref{sec:conclusion} concludes our work and discusses future directions.

\section{Preliminaries}\label{sec:prelim}

We first provide the basic MDP terminologies of language modeling and define the problem formulation of online preference alignment.
\subsection{The Language MDP and RLHF w.r.t. an Individual Preference}\label{sec:language_mdp}

The context of generating a sequence of tokens auto-regressively with an LLM can be modeled as an MDP, where the state ($s_t$) at timestep $t$ is the sequence of tokens generated so far, $s_t = (a_0, a_1, \dots, a_{t-1})$, action ($a_t$) at timestep $t$ is the next token to be generated, chosen from a finite vocabulary $\mathcal{A}$, The state transition is deterministic, i.e., given state $s_t$ and action $a_t$, the next state is simply their concatenation: $s_{t+1} = \text{concat}(s_t, a_t)$, and the policy ($\pi(a_t | s_t)$) is a conditional LLM that we aim to optimize. It gives the probability of generating token $a_t$ given the preceding sequence $s_t$. The goal is to find an optimal policy $\pi$ that solves the following maximum entropy (soft) RL problem: 
\begin{equation}
\max_{\pi} \mathbb{E}_{\pi} \left[ \sum_{t=0}^{T-1} r(s_t, a_t) - \beta D_{KL} (\pi(\cdot | s_t) \| \pi_{\text{ref}}(\cdot | s_t))  \right],\label{eq:rlhf}
\end{equation}
where $r(s_t, a_t)$ is a reward function that scores the quality of generating token $a_t$ after sequence $s_t$. This can be based on human preference feedback or other quality metrics, $\pi_{\text{ref}}$ is a reference LLM, and $D_{KL}(\pi \| \pi_{\text{ref}}) = \sum_{a \in \mathcal{A}} \pi(a|s) \log \frac{\pi(a|s)}{\pi_{\text{ref}}(a|s)}$ is the Kullback-Leibler (KL) divergence that penalizes the policy $\pi$ for deviating too far from the reference model $\pi_{\text{ref}}$, where the temperature parameter $\beta > 0$ controls the strength of this regularization. 

Given the deterministic transitions $s' = (s,a)$, and the fact that the transition terminates at the final time step $T$, the unique fixed-point solution to this maximum entropy RL problem defines the optimal policy $\pi^*$ in terms of the optimal Q-value function, $Q(s,a)$, that satisfies the soft Bellman backup and the optimal value function, $V(s)$, that serves as normalization \citep{nachum2017bridging}:
\begin{align}
\pi^*(a|s) =\,& \pi_{\text{ref}}(a|s) \exp\left(\frac{Q(s,a) - V(s)}{\beta}\right), \quad \forall s,a, \label{eq:soft_policy}\\
Q(s,a) =\,& r(s,a) + V(s'), \label{eq:soft_q}\\
V(s) =\,& \beta \log \sum_{a \in A} \pi_{\text{ref}}(a|s) \exp(\frac{Q(s,a)}{\beta}), \label{eq:soft_v}
\end{align}
where the bellman backup in \eqref{eq:soft_q} corresponds to deterministic transitions. This shows that the logits of the optimal policy are obtained by adding the optimal Q-values to the logits of the reference policy.

\subsection{Online Preference Alignment as a Multi-objective MDP}\label{sec:personalized_language_mdp}

Adapting LLMs to satisfy diverse and often conflicting user preferences presents a complex multi-objective optimization challenge. Unlike in standard RLHF, where the LLM's generation process is treated as a MDP with a single reward, in online preference alignment, in order to represent a distinct set of user preferences, like conciseness or factual accuracy, one considers a \emph{multi-objective MDP} that involves a preference reward vector, $\mathbf{r}(s,a)=(r_1(s,a),\ldots,r_K(s,a)) \in \mathbb{R}^K$, where each of the $K$ components corresponds to a preference dimension. 
Suppose for any arbitrary new users, her preference can be modeled as a linear combination of these multifaceted preference attributes, i.e., $r_{\mathbf{w}}=\sum_{k=1}^{K} w_k r_k$, w.r.t. the user-specific preference vector $\mathbf{w}=(w_1,\ldots,w_K)\in\Delta^K$ that  lies in a $K$-dimensional simplex set, characterizing the underlying user-specific importance placed on the corresponding base rewards. Then a typical approach for learning a tailored LLM is via RLHF \citep{kirk2023personalisation, das2024active}, i.e., $\max_{\pi} \mathbb{E}_{\pi} \left[ \sum_{t=0}^{T-1} r_{\mathbf{w}}(s_t, a_t) - \beta D_{KL} (\pi(\cdot | s_t) \| \pi_{\text{ref}}(\cdot | s_t))  \right]$, which optimizes a policy $\pi^*_\mathbf{w}$ from the feedback signal of a specific reward model $r_{\mathbf{w}}$. However, in general this preference vector $\mathbf{w}$ is not revealed to the agent prior to online interactions. Estimating this vector when concurrently training a corresponding tailored policy via RLHF can be challenging (e.g., beyond policy optimization one may require advanced exploration strategies to uncover such user preferences during RL), especially when this procedure is only run for a limited number of steps (e.g., during online adaptation). Alternatively, training a contextual agent for every possible  preference vector $\mathbf w$ can also be computationally expensive and impractical for most real-world applications.

\section{A Spectral Representation for Online Preference Alignment}\label{sec:universal_rep}
This section investigates the parameterization of the optimal value function within the class of Language MDPs introduced in Section~\ref{sec:language_mdp}. Given the reference LLM representation $\psi(s) \in \mathbb{R}^d$, where the reference policy can be expressed as
$
\pi_{\text{ref}}(a|s) = \exp(\psi(s)^\top \nu_{\text{ref}}(a))/\int_{b \in A} \exp(\psi(s)^\top \nu_{\text{ref}}(b)) db
$ with the corresponding action token embedding $\nu_{\text{ref}}(a) \in \mathbb{R}^d$, our primary goal is to identify the conditions such that this reference LLM feature can also be a \emph{spectral representation} that permits a linear parameterization of the optimal Q-function defined in Equation~\eqref{eq:soft_q}. To facilitate our analysis of the spectral representation for the optimal Q-function, we introduce two technical assumptions that provide a practical framework for our analysis:
\begin{assumption}[Linear Reward Representation]\label{assumption:linear_reward}  Given sufficiently expressive features $\psi$ derived from the reference LLM, any reward function of the language MDP can be linearly represented by these features:
    \begin{equation}
        r(s,a) = \psi((s,a))^\top \nu_{\text{r}}, \quad \text{for some weight vector } \nu_{\text{r}}.
    \end{equation}
    \end{assumption}
    \begin{assumption}[$L$-step Decodability]\label{assumption:l_step_decodability}
          The language MDP induced by the reference LLM is $L$-step decodable for some integer $L>0$, whose trajectory distribution depends only on its most recent $L$-step history, i.e., the distribution of $h$-step trajectory $\tau_h$ is conditioned only on sub-sequence $\tau_{h-L+1:h}=(s_{h-L+1}, a_{h-L+1}, \dots, s_h)$.
    \end{assumption}
The linear reward assumption is justified by the powerful representational capacity of the reference LLM. While the true reward dynamics may be arbitrarily complex, the LLM-generated features $\psi(s)$ are rich enough to represent the underlying semantics, allowing the reward itself to be modeled as a simple linear function. The $L$-step decodability assumption is motivated by the architecture of the transformer-based reference policy. These models operate on a fixed-length context window, meaning their outputs are conditioned only on the most recent $L$ tokens.
Leveraging these conditions that align our  model with the realistic computational constraints of the LLM, we first have our main technical result characterizing the optimal Q-function of any reward function that satisfies Assumption \ref{assumption:linear_reward}.
\begin{lemma}\label{lem:linearly_parameterized_q}
For any language MDP, that satisfies Assumption \ref{assumption:linear_reward} and Assumption \ref{assumption:l_step_decodability}, its optimal Q function from Equation \ref{eq:soft_q} can be linearly parameterized with the reference LLM logit feature $\psi$, i.e., there exists a vector $\nu_{\beta,r,\text{ref}}\in\mathbb R^d$ that depends on temperature $\beta$, reward $r$, and reference LLM $\pi_{\text{ref}}$ such that
\begin{equation}
Q^*(s, a)=\psi((s,a))^\top \nu_{\beta,r,\text{ref}},\,\forall s,a.
\end{equation}
\end{lemma}
Lemma~\ref{lem:linearly_parameterized_q} reveals a non-trivial, crucial property of the language MDP: the reference LLM's logit feature, $\psi$, acts as a universal \emph{spectral representation}. This representation allows the optimal Q-function for any preference-driven reward to be linearly parameterized. Consequently, for a set of $K$ distinct preference attributes $\{r_1(s,a),\ldots,r_K(s,a)\}$, their corresponding optimal Q-functions $\{Q^*_{1}(s,a), \ldots, Q^*_{K}(s,a)\}$ can all be expressed as $Q^*_{k}(s,a)=\psi((s,a))^\top \nu_k$, where each $\nu_k \in \mathbb{R}^d$ is a vector in the spectral space. This insight directly motivates our \emph{Spectral Soup} policy architecture. The model employs a shared LLM feature extractor that produces $\psi$, while integrating multiple lightweight adapters at the output logit layer, each specialized to learn a basis Q-function $Q^*_k$. A souped policy, $\tilde{\pi}_{\lambda}$, is then constructed by linearly combining these Q-functions with a mixture vector $\lambda = (\lambda_1, \ldots, \lambda_K) \in \mathbb{R}^K$:
\begin{equation}\label{eq:spectral_soup_policy}
    \tilde{\pi}_{\lambda}(a|s) \propto \pi_{\text{ref}}(a|s) \cdot \exp\left( \sum_{k=1}^{K} \lambda_k Q^*_k(s,a)/\beta'\right),\,\,\beta\sum_k|\lambda_k| \leq \beta'.
\end{equation}
Here, $\beta' > 0$ acts as a temperature parameter, normalizing the logit mixture's magnitude, and is constrained to maintain temperature consistency with the optimal policies.
Mathematically, when the number of basis functions approximates the spectral dimension ($K \approx d$), parameterizing the personalized policy via the mixture vector $\lambda \in \mathbb{R}^K$ is equivalent to learning a single spectral feature vector $\nu \in \mathbb{R}^d$. However, our approach of mixing Q-functions offers two  advantages. First, the basis of Q-functions is interpretable, as each $Q^*_k$ corresponds to a tangible preference attribute. The underlying spectral representation, in contrast, is often not. This allows us to select a small, relevant subset of basis functions ($K \ll d$) that spans most preference-alignment needs, reducing the learning problem's dimensionality in practice. Second, this framework is flexible; one can easily modify the basis by adding or removing specialized Q-functions to accommodate new preference attributes. Altering the spectral representation, however, is difficult as it is an intrinsic property of the reference LLM.

The central goal is to derive a performance sub-optimality bound for the computationally-efficient spectral soup policy.
To establish this bound, for any given user preference vector $\mathbf{w}$ we first define the optimal target. Recall that a user's preference $\mathbf{w}$ defines a personalized reward, $r_{\mathbf{w}}(s,a) = \sum_{k=1}^{K} w_k r_k(s,a)$. The optimal policy, $\pi^*_{\mathbf{w}}$, that maximizes the KL-regularized return for this reward has the following closed-form solution:
$
    \pi^*_{\mathbf{w}}(a|s) \propto \pi_{\text{ref}}(a|s) \cdot \exp\left({Q^*_{\mathbf{w}}(s, a)}/{\beta}\right),
$
where $Q^*_{\mathbf{w}}$ is the optimal personalized Q-function. The corresponding optimal value function, which represents the maximum performance utility that can be achieved, is $V^*_{\mathbf{w}}(s) = \beta \cdot \log\mathbb{E}_{a\sim \pi_{\text{ref}}(\cdot|s)} \left[\exp\left({Q^*_{\mathbf{w}}(s,a)}/{\beta}\right)\right]$. Our approach approximates this optimal policy with the spectral soup policy $\tilde{\pi}_{\lambda}$ in Equation \eqref{eq:spectral_soup_policy}, whose logit-mixture weights $\lambda^*$ is a solution of the following constrained optimization problem:
\begin{equation}\label{eq:spectral_soup_opt}
    V^{\lambda^*}_{\mathbf{w},\beta'}(s) := \max_{\lambda\in\mathbb{R}^K} \left\{ \mathbb{E}_{\tilde\pi_\lambda} \left[ \sum_{t=0}^{T-1} r_{\mathbf{w}}(s_t, a_t) - \frac{\beta'}{\sum_k|\lambda_k|} D_{KL} (\tilde\pi_\lambda(\cdot | s_t) \| \pi_{\text{ref}}(\cdot | s_t)) \bigg| s_0=s\right] \,\,\,\text{s.t.}\,\,\, \beta\sum_k|\lambda_k|\leq \beta'\right\}.
\end{equation}
This problem maximizes the same preference reward $r_{\mathbf{w}}$ as in RLHF, balanced by a KL penalty whose regularization strength adapts to the magnitude of the policy mixture. Such a key constraint ensures the resulting policy's temperature remains within a reasonable range, which is also a necessary condition for our performance guarantees to hold.
The following theorem presents our main technical result, establishing the formal guarantees for this approximation.

\begin{theorem}[Sub-optimality Performance Bounds]\label{thm:q_soups}
    Under Assumptions \ref{assumption:linear_reward} and \ref{assumption:l_step_decodability}, the spectral soup policy $\tilde{\pi}_{\lambda^*}$ in Equation \eqref{eq:spectral_soup_policy}, whose weights $\lambda^*$ solves Equation \eqref{eq:spectral_soup_opt}, achieves the following guarantees.
    
    \textbf{1. KL Divergence Bound:} The divergence from the true optimal policy $\pi^*_{\mathbf{w}}$ is bounded by:
    \begin{align}\label{eq:kl_bdd}
        D_{\text{KL}}(\pi^*_{\mathbf{w}}(\cdot|s) \,\|\, \tilde{\pi}_{\lambda^*}(\cdot|s))
        \leq \frac{1}{\beta'}\left( \mathbb{E}_{\pi^*_{\mathbf{w}}} \|\psi((s,a))\|_2 + \mathbb{E}_{\pi_{\text{ref}}} \|\psi((s,a))\|_2 \right) \cdot \left\|\frac{\beta'}{\beta}\nu_{\beta,r_{\mathbf w},\text{ref}} - \sum_k\lambda^*_k \nu_{\beta,r_k,\text{ref}}\right\|_2.
    \end{align}
    
    \textbf{2. Performance Sub-optimality Bound:} The gap between the optimal value $V^*_{\mathbf{w}}(s)$ and the value achieved by the spectral soup policy, $V^{\lambda^*}_{\mathbf{w},\beta'}(s)$, is bounded as:
    \begin{equation}
    \begin{split}
       0\leq  V^*_{\mathbf{w}}(s) &- V^{\lambda^*}_{\mathbf{w},\beta'}(s) \leq  \mathbb{E}_{\underline{\pi}}\left[\sum_{t=0}^{T-1}\|\psi((s_t,a_t))\|_2 | s_0=s\right] \cdot \left\|\sum_k\nu_{r_k}\left(w_k-\frac{|\lambda^*_k|}{\sum_k|\lambda^*_k|}\right)\right\|_2 \\
        & + \frac{\beta^2}{\beta'}\sum_k\Delta_k(s)\max\{0,-\lambda^*_k\} + \mathbb{E}_{\pi_{\text{ref}}}\left[\|\psi((s,a))\|_2\right] \cdot \left\|\nu_{\beta,r_{\mathbf w},\text{ref}}-\frac{\beta}{\beta'}\sum_k\lambda^*_k \nu_{\beta,r_k,\text{ref}}\right\|_2.
    \end{split}
    \end{equation}
    Here, $\underline\pi$ is an auxiliary policy that minimizes the weighted difference of cumulative rewards, i.e.,  $\underline\pi\in\arg\min_{\pi}\mathbb E_{\pi}[\sum_{t=0}^{T-1}\psi((s_t,a_t))^\top\sum_k|\lambda^*_k|(\nu_{r_{\mathbf{w}}}-\nu_{r_k})]$, and $\Delta_k(s):=\log (\overline{M}_k(s)+\underline{M}_k(s) - 1) - \log(\overline{M}_k(s)\cdot\underline{M}_k(s))$, where $\overline M_k(s)$ and $\underline M_k(s)$ are the respective upper and lower bounds of the policy ratio $\pi_k(a|s)/\pi_{\text{ref}}(a|s)$.
\end{theorem}
This technical result marks a significant advancement for policy ``souping'' methods, which have largely remained empirically-validated heuristics. Our work establishes one of the first formal sub-optimality bounds for such a technique, providing a rigorous theoretical foundation to complement its practical efficacy. Theorem~\ref{thm:q_soups} provides this guarantee by first bounding the policy approximation error (measured by the KL divergence) between our spectral soup policy and the personalized optimal policy  with the \emph{logit approximation error}, $\|\nu_{\beta,r_{\mathbf{w}},\text{ref}}-({\beta}/{\beta'})\sum_k\lambda^*_k\, \nu_{\beta,r_k,\text{ref}}\|_2$, which captures how well the personalized logit-vector can be represented by a linear combination of the basis vectors (normalized by the magnitude of the logit-mixture weights). Furthermore, it decomposes the performance gap into three intuitive error sources: (i) this same logit approximation error; (ii) a \emph{reward approximation error}, $\|\sum_k\nu_{r_k}(w_k-|\lambda^*_k|/{\sum_k|\lambda^*_k|})\|_2$, capturing the mismatch between the user's true preference weights $\{w_k\}_{k=1}^K$ (which is a simplex vector) and the normalized magnitudes of the learned soup weights $\{\lambda_k\}_{k=1}^K$; and (iii) a \emph{penalty for negative logit-mixture weights}. This decomposition reveals a critical insight: if the set of basis logit-vectors is rich enough to span the spectral representation space of all possible personalized logits, the policy approximation error vanishes entirely. However, the theorem also highlights a limitation, as the overall performance bound is not \emph{tight}; it does not go to zero even when this primary error term is eliminated, due to the remaining reward approximation and penalty terms. This provides a clear direction for future work on tightening these theoretical guarantees.

\begin{remark}[Discovery of Spectral Representation] 
Our core objective is to find a suitable spectral representation $\psi$ that satisfies the above theoretical assumptions, and is amenable to our policy souping methodology. 
A first approach is to learn the representation $\psi$ using an Expectation-Maximization (EM) style algorithm to satisfy Assumption \ref{assumption:linear_reward}. This process alternates between two main steps until convergence: (i) With a fixed set of basis reward vectors $\nu_k$, we first optimize the representation $\psi$ by minimizing the reconstruction error of a weighted reward mixture: $\min_{\psi} \mathbb{E}_{w \sim \Delta_K} \mathbb{E}_{(s,a) \sim \mathcal{D}} \left( r_w - \sum_{k} w_k\cdot \psi(s,a)^\top \nu_{r_k} \right)^2$. This step seeks a representation that allows the $K$ basis vectors to efficiently span the space of personalized rewards; (ii) With a fixed representation $\psi$, we update the basis vectors to best reconstruct each individual reward function by solving:
       $\min_{\nu_1, \dots, \nu_K \in \mathbb{R}^d} \sum_{k} \mathbb{E}_{(s,a) \sim \mathcal{D}} \left[ (r_k - \psi(s,a)^\top \nu_{r_k})^2 \right]$.
Second, as a practical alternative motivated by our value approximation results in Theorem \ref{thm:q_soups}, we propose discovering the spectral representation within a powerful, reference LLM to minimize approximation errors for the mixture of Q-functions. Instead of learning this representation from scratch, we leverage the rich semantic features within the model's intermediate transformer layers (the auto-regressive features). Departing from previous analysis that focuses only the final pre-logit layer, our implementation applies low-rank adapters (LoRA) \citep{wang2023lora} to all intermediate layers when learning the specialized Q-functions $\{Q^*_1, \ldots, Q^*_K\}$. This strategy reframes the problem from representation learning to representation discovery; the algorithm automatically searches for and amplifies features from the most suitable layers by updating their LoRA modules during fine-tuning, allowing the model to identify an optimal spectral representation for each specialized Q-function. Furthermore, the $\lambda$-weight-mixture approach can be extended to the LoRA modules in each of the $L$ intermediate layers. This exponentially expands the number of basis Q-functions from $K$ to $K^L$, creating a powerful mechanism that can significantly reduce the value approximation error.
\end{remark}

\section{The Spectral Souping Algorithm for Online Preference Alignment} \label{sec:alg}
Motivated by the theoretical guarantees in Theorem~\ref{thm:q_soups}, we propose a spectral souping algorithm designed for the rapid and efficient preference alignment of LLMs. The algorithm operates in two phases: an offline training stage followed by an online adaptation stage. In the first phase, we train a set of $K$ specialized policies, $\{\pi_1^*, \dots, \pi_K^*\}$, where each policy is independently optimized for a single reward attribute $r_k$. In the second phase, the logit-mixture vector $\lambda$ of the spectral soup policy $\tilde{\pi}_{\lambda}$ in Equation~\eqref{eq:spectral_soup_policy} is learned online to combine the specialized policies, tailoring the model's behavior to a specific user's preferences. This structured composition enables zero-shot generalization to new users by learning only the low-dimensional mixture vector $\lambda$ at inference time, rather than training a bespoke policy for each user's underlying preference vector $\mathbf{w}$.

The initial phase involves learning a set of $K$ base policies $\{\pi^*_k\}_{k=1}^K$. Within the language MDP framework (Section~\ref{sec:language_mdp}), each specialized policy is given by $\pi^*_{\theta_k}(a|s) \propto \pi_{\text{ref}}(a|s)\exp(Q_{\theta_k}(s,a)/\beta)$, implying its output logits can be expressed as $
\text{logit}_{\theta_k}(s,a) = \text{logit}_{\text{ref}}(s,a) + Q_{\theta_k}(s,a)$. Following spectral representation theory, the Q-function $Q_{\theta_k}(s,a)$ can be optimally parameterized using a LoRA module applied to the output logit layer of the reference policy $\pi_{\text{ref}}$.
A practical challenge is the lack of granular, token-level rewards. We address this by learning from $K$ offline datasets, $\{\mathcal{B}_k\}_{k=1}^K$, where each dataset contains trajectories annotated with binary labels (e.g., "good" vs. "bad") for a specific preference attribute. This reframes the task of learning each specialized policy's parameters, $\theta_k$, as a well-studied offline preference alignment problem.
Let $R_{\theta_k}(\tau):= \sum_{t=0}^{T-1}r(s_t,a_t)=\sum_{t=0}^{T-1} Q_{\theta_k}(s_t, a_t) - \sum_{t=1}^{T}V_{\theta_k}(s_{t})$ be the cumulative reward over trajectory $\tau$, derived according to Equation \eqref{eq:soft_q}.
One method, developed from \citet{cui2025process}, explicitly learns the Q-function, $Q_{\theta_k}(s,a)$, by minimizing a composite objective. This objective combines a binary loss, i.e., $L_{\text{Binary}} = -\mathbb{E}_{\tau \sim \mathcal{B}_k} [l_k(\tau)\log\sigma(R_{\theta_k}(\tau)) + (1-l_k(\tau))\log(1-\sigma(R_{\theta_k}(\tau)))]$, to match trajectory labels with a Gumbel loss \citep{garg2023extreme}, i.e., $L_{\text{Gumbel}} = \mathbb{E}_{(s,a) \sim \mathcal{B}_k} \left[ \exp({A_{\theta_k}(s,a)}/{\beta}) - {A_{\theta_k}(s,a)}/{\beta} - 1 \right]$, where $A_{\theta_k}(s,a) = Q_{\theta_k}(s,a) - V_{\theta_k}(s)$, to ensure soft Bellman consistency.
While direct, this approach is computationally expensive because it requires training an auxiliary value model, $V_{\theta_k}$. An alternative method, inspired by \citet{rafailov2024r}, operates on preference pairs $(w,l)$ derived from $\mathcal{B}_k$, where trajectory $w$ is preferred over trajectory $l$. It optimizes the policy by minimizing the Bradley-Terry logistic loss:
$\mathcal{L}_{BT}(\theta_k; \mathcal{B}_k) = -\mathbb{E}_{(w,l) \sim \mathcal{B}_k}[\log\sigma(R_{\theta_k}(w) - R_{\theta_k}(l))]$.
The key advantage of this formulation is that the difference in cumulative rewards simplifies to a sum of log-policy ratios, i.e., $R_{\theta_k}(w) - R_{\theta_k}(l) = \beta\sum_t \log {\pi^*_{\theta_k}(a_{w,t}|s_{w,t})}/{\pi_{\text{ref}}(a_{w,t}|s_{w,t})} - \beta\sum_t \log {\pi^*_{\theta_k}(a_{l,t}|s_{l,t})}/{\pi_{\text{ref}}(a_{l,t}|s_{l,t})}$, thereby obviating the need for an explicit value function. Although this determines the reward function only up to a state-dependent potential, this ambiguity does not affect the optimal policy. For its computational efficiency, we adopt this second method in our work.

The second phase is online adaptation, via spectral souping, where the policy can be realized through two approaches.
The \emph{explicit} approach directly constructs the policy's logit function as a linear combination of the pre-trained specialized and reference logits:
$
\text{logit}_{\tilde\pi_\lambda}(s,a) = \left(1-\sum_k \lambda_k\right)\text{logit}_{\text{ref}}(s,a) + \sum_k \lambda_k\text{logit}_{\theta_k}(s,a).
$
This follows directly from the definition of the specialized Q-functions and the spectral soup policy in Equation~\eqref{eq:spectral_soup_policy}.
Alternatively, the \emph{implicit} approach avoids instantiating a new model by applying rejection sampling to the reference policy $\pi_{\text{ref}}$. An action is first sampled from the reference policy, $a \sim \pi_{\text{ref}}(a|s)$, and then accepted if $u \le \exp(\sum_{k=1}^K \frac{\lambda_k}{\beta}(\log \pi^*_{\theta_k}(a|s) - \log\pi_{\text{ref}}(a|s)))$, 
where $u$ is drawn from a uniform distribution $\mathcal{U}[0, M(s)]$, and the upper bound can be simply set to $M(s)=\exp(-\sum_{k} \lambda_k\,D_{KL}(\pi_{\text{ref}}(\cdot|s)||\pi^*_{\theta_k}(\cdot|s))/\beta)$. While this implicit souping method can be highly efficient, its performance degrades if the personalized policy diverges significantly from the reference policy, as this leads to a high rejection rate. For both approaches, the spectral soup weights $\lambda \in \mathbb{R}^K$, which tailor the policy to a specific user, are learned efficiently from online user feedback. Mirroring the offline phase, we can frame the learning problem using preference optimization, where we minimize a Bradley-Terry loss over weighted preference pairs, $\mathcal{L}_{\text{BT}}(\lambda; \mathcal{B}) = -\mathbb{E}_{ \mathcal{B}} [\log\sigma(\sum_k\lambda_k(R_{\theta_k}(w) - R_{\theta_k}(l)))]$. Crucially, the learning problem for $\lambda$ reduces to  logistic linear regression, allowing for efficient update via convex optimization methods \citep{hazan2016introduction}.

\section{Experiments}\label{sec:exp}
To assess our approach's effectiveness, we conduct empirical evaluations on three realistic LLM online preference alignment benchmarks. Each experiment involves an offline phase for learning specialized policies and an online phase for personalization and adaptation.

Our first experiment is on online LLM preference alignment w.r.t. the \textbf{UltraFeedback} dataset \citep{cui2023ultrafeedback}, which contains prompts with response pairs annotated with a 4D feature vector (helpfulness, honesty, instruction-following, and truthfulness). We synthesize diverse preferences by ranking responses based on the dot product of their feature vector with a unique weight vector $\mathbf{w} \in \mathbb{R}^4$. For the offline phase, we build $K=30$ specialized datasets by creating 30 unique weight vectors, each randomly sampled around the basis attributes. In the online phase, we test generalization against 5 held-out users proxied by unseen public reward models. To increase difficulty, the dataset is filtered to only contentious examples ($23,614$ train, $401$ test) where preferences conflict. The second experiment focuses on optimizing an LLM's personalized prompt-expansion for \textbf{text-to-image (T2I) generation} in a 5-turn interactive process \citep{nabati2024preference}, where an agent generates new textual prompts at each turn and inputs that to the environment to return the updated 4x4 image slate. The system uses Stable Diffusion XL \citep{podell2023sdxl} for image generation, Gemini 1.5 Flash \citep{team2024gemini} for prompt expansion, and Gemma 2B \citep{team2024gemma} to model utilities. In the offline phase, we generate $K=32$ datasets from over 30,000 simulated rollouts guided by user models with myopic, turn-by-turn preferences; a user rates best overall image in each column and chooses the highest-scoring column. During the online phase, adaptability is tested against 5 held-out users, each simulated by a pre-trained LTV utility function that models session-level preferences based on criteria like aesthetics or prompt consistency. Our third experiment is on LLM \textbf{sleep coaching}. Each synthetic user is instantiated by an LLM grounded in their detailed sleep profiles obtained from 68 real individuals in the LifeSnaps dataset \citep{Yfantidou2022-ay}, following the experimental setup in \citet{yun-etal-2025-sleepless}. For the offline phase, we generate three (high, medium, low) preference datasets for each of the Big Five personality dimensions \citep{goldberg1992development} (the International Personality Item Pool (IPIP) version; extraversion, agreeableness, conscientiousness, stability, intellect), resulting in $K=15$ personality types. Each dataset contains 1,000 10-turn conversation pairs, ranked by a reward function specific to that personality. In the online phase, we evaluate performance on 5 simulated users (512 samples each), each represented by a Gemini 1.5 Flash auto-rater that scores conversations based on its specialized rubrics on coaching tone, user understanding, and intervention quality, that mimics users of different backgrounds and personalities.

To evaluate the performance of our spectral souping method, we compare it against a comprehensive suite of baselines. The first is the ``bespoke'' RLHF \citep{ouyang2022training} agent, which is trained directly on a specific user's feedback. This approach serves as a practical upper bound for performance but is computationally expensive and does not generalize across different users. To tackle the multi-objective nature of the task, Personalized Soups (P-SOUPS) \citep{jang2023personalized} trains multiple specialized policies, each optimized for a single objective. At inference time, it creates a personalized policy by mixing the model parameters of these specialized policies via a heuristic weighted average. This technique is analogous to our explicit spectral souping approach. We also compare against two decoding-time alignment frameworks, which are conceptually similar to our implicit spectral souping approach. The Personalized Alignment at Decoding-time (PAD) framework \citep{chen2024pad} guides the decoding process by mixing the model's output distribution based on a preference reward vector. Inspired by successor features, the PAD-SF variant \citep{barreto2018transfer} re-weights the output distribution using a softmax distribution derived from a learnable K-dimensional scorer, which operates on a spectral representation, mimics the specialized advantage functions. For our experiments, the default souping models are based on the Gemma-V3 \citep{team2025gemma} 4B, 1B, 270M architectures, implemented with rank-100 LoRA modules on all intermediate layers.

\begin{figure}[htbp]
    \centering
    \includegraphics[width=\textwidth]{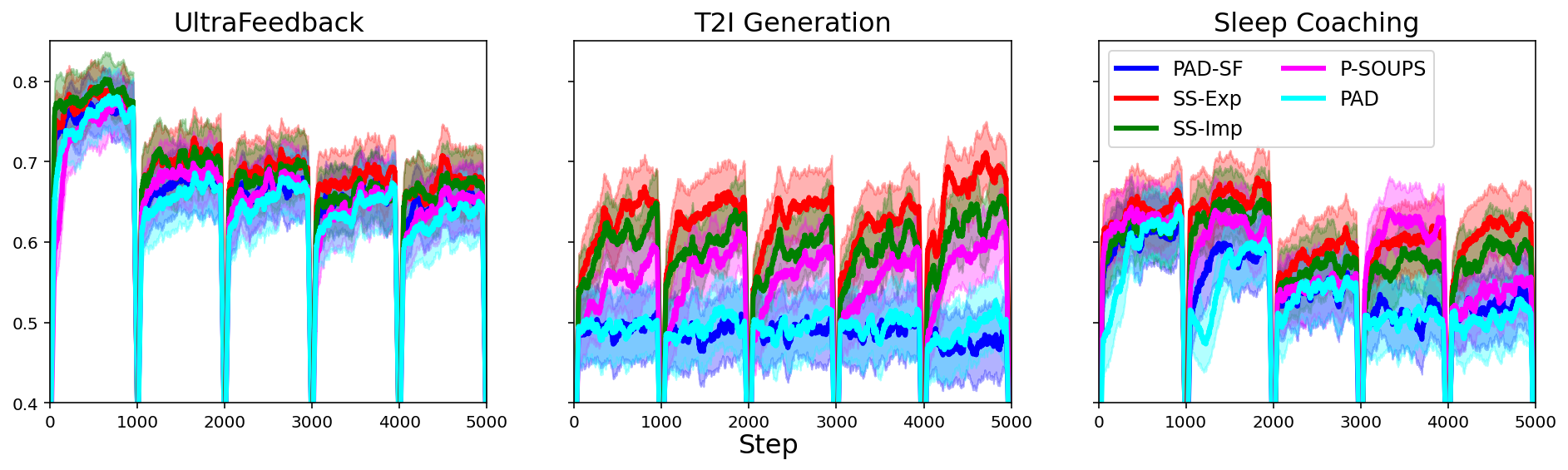} 
    \caption{Test-time Training Performance of Different Methods: Explicit \& Implicit Spectral Souping (SS-Exp \& SS-Imp), P-SOUPS, PAD, PAD-SF, RLHF, adapted to 5 various users in the UltraFeedback, T2I Generation, Sleep Coaching domains. The SS methods (especially SS-Exp) consistently and outperform P-SOUPS and the PAD baselines, demonstrating superior performance in online adaptation.}
    \label{fig:test_time_training_4B}
\end{figure}

\begin{figure}[htbp]
    \centering
    \includegraphics[width=\textwidth]{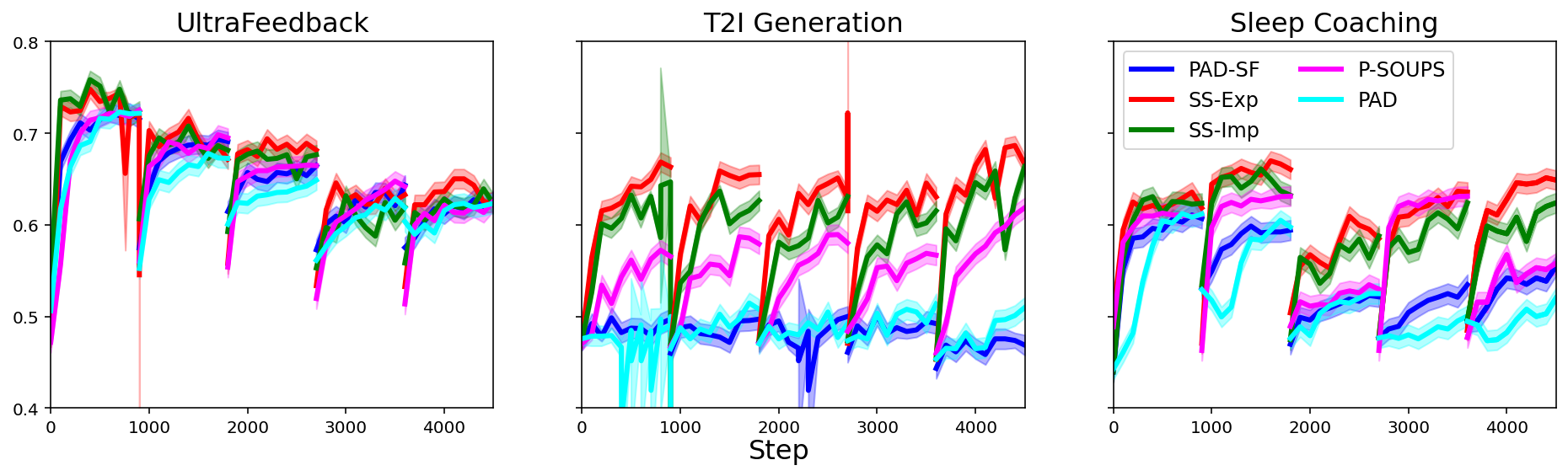} 
    \caption{Evaluation Performance of Different Online Adaptation Methods: Explicit \& Implicit Spectral Souping (SS-Exp \& SS-Imp), P-SOUPS, PAD, PAD-SF, RLHF, across adapted to 5 various users in the UltraFeedback, T2I Generation, Sleep Coaching domains. The superior performance of the SS methods (over P-SOUPS, PAD, and PAD-SF baselines) is also generalizable to online evaluations.}
    \label{fig:test_time_eval_4B}
\end{figure}

Our experimental results, summarized in Figures \ref{fig:test_time_training_4B} \& \ref{fig:test_time_eval_4B} for GemmaV3 4B and Figures \ref{fig:test_time_training_1B} \& \ref{fig:test_time_eval_1B} for GemmaV3 1B (in Appendix \ref{appendix:results}), validate the efficacy of spectral souping across different domains. A key observation is that implicit and explicit spectral souping methods achieve comparable performance, with the explicit approach holding a slight edge in several settings. This aligns with our hypothesis that both methods stem from the same theoretical foundation described in Section \ref{sec:universal_rep}, but explicit souping is more precise as it avoids the sampling approximations inherent to the implicit method. Crucially, our methods consistently approximate the performance of the computationally expensive, tailored RLHF upper bound, when averaged over all online users, achieving $83\%$ of the optimal performance on UltraFeedback, $88\%$ on T2I, and $72\%$ on Sleep coaching, thereby empirically confirming our theoretical sub-optimality bounds. When compared against other baselines, spectral souping demonstrates significant advantages. It surpasses P-SOUPS, even after tuning the hyper-parameters (the P-SOUPS weights) used for averaging, underscoring the necessity of merging policies within the structured spectral representation space rather than an arbitrary parameter space. Similarly, implicit spectral souping is more stable and performs better than PAD and PAD-SF baselines. This suggests that guiding the policy mixture with optimal Q-function weights in the spectral representation space is more effective than using reward weights (as in PAD) or souping weights heuristically developed via advantage approximations with successor features (as in PAD-SF). Finally, across all held-out users in both T2I and sleep coaching, spectral souping exhibits faster and more robust online adaptation, proven to be more data-efficient than all baselines, highlighting its capability  in real-world scenarios. 

\begin{figure}[htbp]
    \centering
    \includegraphics[width=\textwidth]{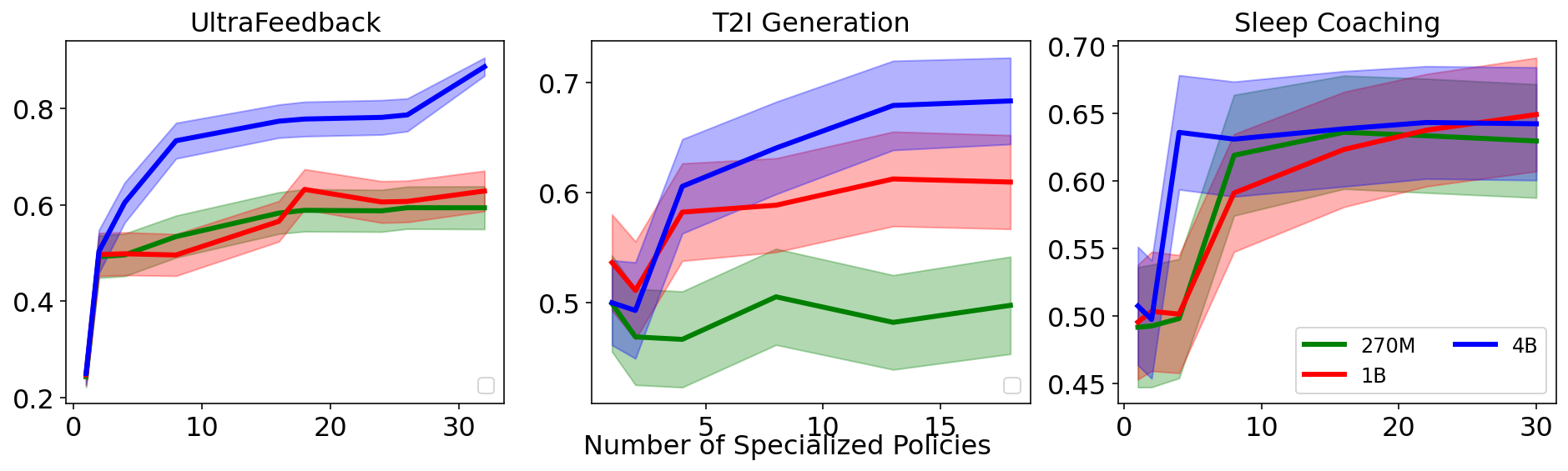} 
    \caption{Scaling Laws of SS-Exp to illustrate the effect of model size on performance with an increasing number of specialized policies across three  domains. While performance consistently improved with model size, the larger (4B) model architecture developed a more comprehensive spectral representation, making its online adaptation agents more robust to basis reductions.}
    \label{fig:test_time_ablation}
\end{figure}

Our ablation study investigates the impact of reducing the number of offline specialized policies ($K$) on online adaptation performance, with results presented in the scaling-law curves in Figure \ref{fig:test_time_ablation} with analysis conducted across our 3 domains using the Gemma-V3 4B, 1B and 270M models respectively. To manage the combinatorial complexity of basis selection of specialized policies, we employed a leave-one-out elimination approach, randomly selecting and removing one specialized policy at a time, and averaging the results. The findings reveal several key insights. First, as illustrated in both figures, reducing the number of basis policies consistently degrades online learning performance across all domains, as expected. More importantly, we observe a significant performance drop below a certain threshold for each domain—specifically at $K=7$ for UltraFeedback, $K=5$ for T2I generation, and $K=13$ for sleep coaching. This suggests the existence of a minimal set of specialized policies is required to form a basis that effectively span the representation space for various preferences. Second, the size of this minimal basis set appears correlated with the complexity of the problem's underlying characteristics; domains like UltraFeedback (4 core attributes) and T2I (5 preference categories) require fewer basis policies than the real-world sleep coaching domain, which involves more diverse and complex auto-rater feedback. Finally, the larger Gemma-V3 4B model exhibits a more gradual decrease in performance as specialized policies are removed. This suggests that larger pretrained models capture a more semantically comprehensive spectral representation, making each specialized policy more expressive and the overall basis more robust to reductions in its span.

\section{Related Work}\label{sec:related_work}
Our work is situated at the intersection of LLM preference alignment, RL spectral representation, and online adaptation. We review the key concepts in these areas that relate to our methods below.

\paragraph{LLM Alignment}{A dominant approach to aligning LLMs with human preferences is RLHF \citep{christiano2017deep, bakker2022fine}, which involves training a reward model to mimic human feedback and then using an algorithm like Proximal Policy Optimization (PPO) \citep{schulman2017proximal} to fine-tune the LLM. Direct Policy Optimization (DPO) has also emerged as a promising alternative that simplifies the above process and reduces computational cost. It reformulates the RLHF problem as a preference-based classification task, optimizing a policy without the need for an explicit reward model. More recently, research has explored decoding-time alignment as a way to avoid the computational cost of model training. This includes methods like Controlled Decoding (CD) \citep{mudgal2023controlled}, which uses a value-based scorer to guide generation, ARGS \citep{khanov2024args}, which adjusts probabilistic predictions based on online feedback, and DeAL \citep{huang2024deal} which focuses on heuristic-guided searches to meet diverse alignment goals. While these methods are powerful, they often address alignment w.r.t. a single, uniform preference. Our work extends these concepts by focusing on personalized alignment, which considers the diverse preferences of individual users.}

\paragraph{Spectral Representations for RL}{Representation learning is crucial in RL for abstracting complex state and action spaces to facilitate policy optimization. Existing methods for learning these representations utilize various techniques, including reconstruction \citep{watter2015embed, hafner2019dream, fujimoto2023sale}, successor features \citep{gershman2012successor, kulkarni2016deep, barreto2017successor}, and bisimulation \citep{ferns2014bisimulation, gelada2019deepmdp, zhang2025revisiting}. A particularly effective approach involves using spectral decomposition, which has been explored in works like \citet{mahadevan2007proto} and \citet{ren2022spectral}. These methods often assume that the environment's transition kernel has a low-rank spectral structure, enabling the use of linear representations for value functions and leads to provably sample-efficient algorithms \citep{jin2019bayesian, yang2020reinforcement}. Leveraging the above theoretical underpinnings, our work develops a universal spectral representation in LLMs, whose existence within language MDPs has been proven. This finding justifies our online adaptation algorithm for personalized preference alignment, which is designed to only modify the projection vectors within this stable spectral representation, rather than fine-tuning the entire model. Our approach is distinct from prior work as it connects spectral representation principles to LLM personalization, a domain where such techniques have not been previously explored in this manner.}

\paragraph{Online Preference Alignment}{
Aligning LLMs with online preferences is crucial to addresses widely-varying tastes of individual users \citep{kirk2023past, feng2024modular, jiang2024can, zhang2024self, zhong2024panacea, wang2024arithmetic}. Existing personalized alignment methods can be grouped into three categories: (i) Joint optimization of multi-dimensional preference rewards \citep{zeng2023diversified, li2024dissecting, wang2024arithmetic, zhu2025structured, yang2024aligning, chakraborty2024maxmin, das2024active, zhong2024panacea}; (ii) Merging model parameters (via linearly interpolation) or outputs (via mixture-of-expert composition) w.r.t. multiple preference dimensions \citep{jang2023personalized, rame2024warp, park2024learning, yang2024model, wan2024knowledge}; (iii) Prompt-based methods that use diverse prompts to guide the model toward specific preferences \citep{jafari2024morl, trivedi2025align, hwang2023aligning, min2025prompting, ravichandran2025align}, but that requires textual preference descriptions. Our work combines the first two approaches, which formulates the personalized preference alignment as a Multi-objective language MDP, utilizes its universal spectral representation for LLMs, and develops a theoretically-grounded online adaptation recipe that unifies both model parameter merging and mixture-of-expert output sampling to effectively handle different trade-offs on conflicting preferences.
}

\section{Conclusion}\label{sec:conclusion}
This work introduces spectral souping, a principled framework that addresses the critical challenge of LLM adaptation by discovering a universal spectral representation within the language MDP. This reveals that diverse policies inhabit a structured, low-dimensional latent space, allowing any specialized policy to be effectively represented as a linear combination of a few pre-trained basis policies. This discovery unlocks an efficient two-phase methodology: an offline phase to train a compact set of basis policies, followed by an online inference phase where they are dynamically combined to tailor responses to any user without costly retraining. The efficacy of this approach is validated across several realistic LLM preference-alignment domains where spectral souping consistently outperforms state-of-the-art baselines. For instance, it is significantly more data-efficient than two-stage methods that first infer a tailored reward before applying RLHF, and more stable than alternatives that estimates the occupancy measures (via successor features) of tailored policies. Crucially, unlike prior heuristic approaches, our framework is grounded in formal guarantees, providing sub-optimality bounds that ensure its performance approximates that of a fully fine-tuned LLM. Spectral souping thus offers a scalable, computationally efficient, and theoretically sound solution for online adaptation of LLMs.

This work opens several promising avenues for future research. One key direction is to learn an optimal spectral representation directly during pre-training, embedding a universal basis for diverse human preferences into the foundation model itself to make it inherently more adaptable. Another avenue involves analyzing more sophisticated online adaptation algorithms, such as non-linear souping techniques or meta-learning approaches, to infer user needs more rapidly from sparse feedback. Finally, extending the principles of spectral souping to other modalities (beyond LLMs), including personalized text-to-image generation, presents an exciting opportunity to generalize this framework, enabling robust personalization across a wide range of creative and practical domains.

\bibliography{main}

@article{zhang2023provable,
  title={Provable Representation with Efficient Planning for Partial Observable Reinforcement Learning},
  author={Zhang, Hongming and Ren, Tongzheng and Xiao, Chenjun and Schuurmans, Dale and Dai, Bo},
  journal={arXiv preprint arXiv:2311.12244},
  year={2023}
}

@inproceedings{hau2023entropic,
  title={Entropic risk optimization in discounted MDPs},
  author={Hau, Jia Lin and Petrik, Marek and Ghavamzadeh, Mohammad},
  booktitle={International Conference on Artificial Intelligence and Statistics},
  pages={47--76},
  year={2023},
  organization={PMLR}
}

@article{kulkarni2016deep,
  title={Deep successor reinforcement learning},
  author={Kulkarni, Tejas D and Saeedi, Ardavan and Gautam, Simanta and Gershman, Samuel J},
  journal={arXiv preprint arXiv:1606.02396},
  year={2016}
}

@article{zhang2025revisiting,
  title={Revisiting Bisimulation Metric for Robust Representations in Reinforcement Learning},
  author={Zhang, Leiji and Wang, Zeyu and Li, Xin and Li, Yao-Hui},
  journal={arXiv preprint arXiv:2507.18519},
  year={2025}
}

@article{mahadevan2007proto,
  title={Proto-value Functions: A Laplacian Framework for Learning Representation and Control in Markov Decision Processes.},
  author={Mahadevan, Sridhar and Maggioni, Mauro},
  journal={Journal of Machine Learning Research},
  volume={8},
  number={10},
  year={2007}
}

@article{ren2022spectral,
  title={Spectral decomposition representation for reinforcement learning},
  author={Ren, Tongzheng and Zhang, Tianjun and Lee, Lisa and Gonzalez, Joseph E and Schuurmans, Dale and Dai, Bo},
  journal={arXiv preprint arXiv:2208.09515},
  year={2022}
}

@article{kirk2023past,
  title={The past, present and better future of feedback learning in large language models for subjective human preferences and values},
  author={Kirk, Hannah Rose and Bean, Andrew M and Vidgen, Bertie and R{\"o}ttger, Paul and Hale, Scott A},
  journal={arXiv preprint arXiv:2310.07629},
  year={2023}
}

@article{feng2024modular,
  title={Modular pluralism: Pluralistic alignment via multi-llm collaboration},
  author={Feng, Shangbin and Sorensen, Taylor and Liu, Yuhan and Fisher, Jillian and Park, Chan Young and Choi, Yejin and Tsvetkov, Yulia},
  journal={arXiv preprint arXiv:2406.15951},
  year={2024}
}

@article{jiang2024can,
  title={Can language models reason about individualistic human values and preferences?},
  author={Jiang, Liwei and Sorensen, Taylor and Levine, Sydney and Choi, Yejin},
  journal={arXiv preprint arXiv:2410.03868},
  year={2024}
}

@article{zhang2024self,
  title={Self-exploring language models: Active preference elicitation for online alignment},
  author={Zhang, Shenao and Yu, Donghan and Sharma, Hiteshi and Zhong, Han and Liu, Zhihan and Yang, Ziyi and Wang, Shuohang and Hassan, Hany and Wang, Zhaoran},
  journal={arXiv preprint arXiv:2405.19332},
  year={2024}
}

@article{wang2024arithmetic,
  title={Arithmetic control of llms for diverse user preferences: Directional preference alignment with multi-objective rewards},
  author={Wang, Haoxiang and Lin, Yong and Xiong, Wei and Yang, Rui and Diao, Shizhe and Qiu, Shuang and Zhao, Han and Zhang, Tong},
  journal={arXiv preprint arXiv:2402.18571},
  year={2024}
}

@article{jafari2024morl,
  title={Morl-prompt: An empirical analysis of multi-objective reinforcement learning for discrete prompt optimization},
  author={Jafari, Yasaman and Mekala, Dheeraj and Yu, Rose and Berg-Kirkpatrick, Taylor},
  journal={arXiv preprint arXiv:2402.11711},
  year={2024}
}

@inproceedings{trivedi2025align,
  title={Align-pro: A principled approach to prompt optimization for llm alignment},
  author={Trivedi, Prashant and Chakraborty, Souradip and Reddy, Avinash and Aggarwal, Vaneet and Bedi, Amrit Singh and Atia, George K},
  booktitle={Proceedings of the AAAI Conference on Artificial Intelligence},
  volume={39},
  number={26},
  pages={27653--27661},
  year={2025}
}

@article{hwang2023aligning,
  title={Aligning language models to user opinions},
  author={Hwang, EunJeong and Majumder, Bodhisattwa Prasad and Tandon, Niket},
  journal={arXiv preprint arXiv:2305.14929},
  year={2023}
}

@article{jang2023personalized,
  title={Personalized soups: Personalized large language model alignment via post-hoc parameter merging},
  author={Jang, Joel and Kim, Seungone and Lin, Bill Yuchen and Wang, Yizhong and Hessel, Jack and Zettlemoyer, Luke and Hajishirzi, Hannaneh and Choi, Yejin and Ammanabrolu, Prithviraj},
  journal={arXiv preprint arXiv:2310.11564},
  year={2023}
}

@article{rafailov2023direct,
  title={Direct preference optimization: Your language model is secretly a reward model},
  author={Rafailov, Rafael and Sharma, Archit and Mitchell, Eric and Manning, Christopher D and Ermon, Stefano and Finn, Chelsea},
  journal={Advances in neural information processing systems},
  volume={36},
  pages={53728--53741},
  year={2023}
}

@article{kirk2023personalisation,
  title={Personalisation within bounds: A risk taxonomy and policy framework for the alignment of large language models with personalised feedback},
  author={Kirk, Hannah Rose and Vidgen, Bertie and R{\"o}ttger, Paul and Hale, Scott A},
  journal={arXiv preprint arXiv:2303.05453},
  year={2023}
}

@article{rame2024warp,
  title={Warp: On the benefits of weight averaged rewarded policies},
  author={Ram{\'e}, Alexandre and Ferret, Johan and Vieillard, Nino and Dadashi, Robert and Hussenot, L{\'e}onard and Cedoz, Pierre-Louis and Sessa, Pier Giuseppe and Girgin, Sertan and Douillard, Arthur and Bachem, Olivier},
  journal={arXiv preprint arXiv:2406.16768},
  year={2024}
}

@article{chen2024pad,
  title={Pad: Personalized alignment of llms at decoding-time},
  author={Chen, Ruizhe and Zhang, Xiaotian and Luo, Meng and Chai, Wenhao and Liu, Zuozhu},
  journal={arXiv preprint arXiv:2410.04070},
  year={2024}
}

@article{team2025gemma,
  title={Gemma 3 technical report},
  author={Team, Gemma and Kamath, Aishwarya and Ferret, Johan and Pathak, Shreya and Vieillard, Nino and Merhej, Ramona and Perrin, Sarah and Matejovicova, Tatiana and Ram{\'e}, Alexandre and Rivi{\`e}re, Morgane and others},
  journal={arXiv preprint arXiv:2503.19786},
  year={2025}
}

@article{team2024gemma,
  title={Gemma 2: Improving open language models at a practical size},
  author={Team, Gemma and Riviere, Morgane and Pathak, Shreya and Sessa, Pier Giuseppe and Hardin, Cassidy and Bhupatiraju, Surya and Hussenot, L{\'e}onard and Mesnard, Thomas and Shahriari, Bobak and Ram{\'e}, Alexandre and others},
  journal={arXiv preprint arXiv:2408.00118},
  year={2024}
}

@article{team2024gemini,
  title={Gemini 1.5: Unlocking multimodal understanding across millions of tokens of context},
  author={Team, Gemini and Georgiev, Petko and Lei, Ving Ian and Burnell, Ryan and Bai, Libin and Gulati, Anmol and Tanzer, Garrett and Vincent, Damien and Pan, Zhufeng and Wang, Shibo and others},
  journal={arXiv preprint arXiv:2403.05530},
  year={2024}
}

@article{podell2023sdxl,
  title={Sdxl: Improving latent diffusion models for high-resolution image synthesis},
  author={Podell, Dustin and English, Zion and Lacey, Kyle and Blattmann, Andreas and Dockhorn, Tim and M{\"u}ller, Jonas and Penna, Joe and Rombach, Robin},
  journal={arXiv preprint arXiv:2307.01952},
  year={2023}
}

@article{cui2023ultrafeedback,
  title={Ultrafeedback: Boosting language models with high-quality feedback},
  author={Cui, Ganqu and Yuan, Lifan and Ding, Ning and Yao, Guanming and Zhu, Wei and Ni, Yuan and Xie, Guotong and Liu, Zhiyuan and Sun, Maosong},
  year={2023}
}

@article{nabati2024preference,
  title={Preference Adaptive and Sequential Text-to-Image Generation},
  author={Nabati, Ofir and Tennenholtz, Guy and Hsu, ChihWei and Ryu, Moonkyung and Ramachandran, Deepak and Chow, Yinlam and Li, Xiang and Boutilier, Craig},
  journal={arXiv preprint arXiv:2412.10419},
  year={2024}
}

@article{park2024learning,
  title={Learning more generalized experts by merging experts in mixture-of-experts},
  author={Park, Sejik},
  journal={arXiv preprint arXiv:2405.11530},
  year={2024}
}

@article{yang2024model,
  title={Model merging in llms, mllms, and beyond: Methods, theories, applications and opportunities},
  author={Yang, Enneng and Shen, Li and Guo, Guibing and Wang, Xingwei and Cao, Xiaochun and Zhang, Jie and Tao, Dacheng},
  journal={arXiv preprint arXiv:2408.07666},
  year={2024}
}

@article{wan2024knowledge,
  title={Knowledge fusion of large language models},
  author={Wan, Fanqi and Huang, Xinting and Cai, Deng and Quan, Xiaojun and Bi, Wei and Shi, Shuming},
  journal={arXiv preprint arXiv:2401.10491},
  year={2024}
}

@article{ouyang2022training,
  title={Training language models to follow instructions with human feedback},
  author={Ouyang, Long and Wu, Jeffrey and Jiang, Xu and Almeida, Diogo and Wainwright, Carroll and Mishkin, Pamela and Zhang, Chong and Agarwal, Sandhini and Slama, Katarina and Ray, Alex and others},
  journal={Advances in neural information processing systems},
  volume={35},
  pages={27730--27744},
  year={2022}
}

@article{nachum2017bridging,
  title={Bridging the gap between value and policy based reinforcement learning},
  author={Nachum, Ofir and Norouzi, Mohammad and Xu, Kelvin and Schuurmans, Dale},
  journal={Advances in neural information processing systems},
  volume={30},
  year={2017}
}

@article{min2025prompting,
  title={From prompting to alignment: A generative framework for query recommendation},
  author={Min, Erxue and Huang, Hsiu-Yuan and Yang, Xihong and Yang, Min and Jia, Xin and Wu, Yunfang and Cai, Hengyi and Wang, Junfeng and Wang, Shuaiqiang and Yin, Dawei},
  journal={arXiv preprint arXiv:2504.10208},
  year={2025}
}

@article{ravichandran2025align,
  title={ALIGN: Prompt-based Attribute Alignment for Reliable, Responsible, and Personalized LLM-based Decision-Making},
  author={Ravichandran, Bharadwaj and Joy, David and Elliott, Paul and Hu, Brian and Adams, Jadie and Funk, Christopher and Veenhuis, Emily and Hoogs, Anthony and Basharat, Arslan},
  journal={arXiv preprint arXiv:2507.09037},
  year={2025}
}

@article{zhong2024panacea,
  title={Panacea: Pareto alignment via preference adaptation for llms},
  author={Zhong, Yifan and Ma, Chengdong and Zhang, Xiaoyuan and Yang, Ziran and Chen, Haojun and Zhang, Qingfu and Qi, Siyuan and Yang, Yaodong},
  journal={Advances in Neural Information Processing Systems},
  volume={37},
  pages={75522--75558},
  year={2024}
}

@article{jin2019bayesian,
  title={Bayesian symbolic regression},
  author={Jin, Ying and Fu, Weilin and Kang, Jian and Guo, Jiadong and Guo, Jian},
  journal={arXiv preprint arXiv:1910.08892},
  year={2019}
}

@inproceedings{yang2020reinforcement,
  title={Reinforcement learning in feature space: Matrix bandit, kernels, and regret bound},
  author={Yang, Lin and Wang, Mengdi},
  booktitle={International Conference on Machine Learning},
  pages={10746--10756},
  year={2020},
  organization={PMLR}
}

@inproceedings{gelada2019deepmdp,
  title={Deepmdp: Learning continuous latent space models for representation learning},
  author={Gelada, Carles and Kumar, Saurabh and Buckman, Jacob and Nachum, Ofir and Bellemare, Marc G},
  booktitle={International conference on machine learning},
  pages={2170--2179},
  year={2019},
  organization={PMLR}
}

@inproceedings{ferns2014bisimulation,
  title={Bisimulation Metrics are Optimal Value Functions.},
  author={Ferns, Norman and Precup, Doina},
  booktitle={UAI},
  pages={210--219},
  year={2014}
}

@article{zeng2023diversified,
  title={On diversified preferences of large language model alignment},
  author={Zeng, Dun and Dai, Yong and Cheng, Pengyu and Wang, Longyue and Hu, Tianhao and Chen, Wanshun and Du, Nan and Xu, Zenglin},
  journal={arXiv preprint arXiv:2312.07401},
  year={2023}
}

@article{li2024dissecting,
  title={Dissecting human and llm preferences},
  author={Li, Junlong and Zhou, Fan and Sun, Shichao and Zhang, Yikai and Zhao, Hai and Liu, Pengfei},
  journal={arXiv preprint arXiv:2402.11296},
  year={2024}
}

@article{zhu2025structured,
  title={Structured preference modeling for reinforcement learning-based fine-tuning of large models},
  author={Zhu, Lin and Guo, Fan and Cai, Guohui and Ma, Yumeng},
  journal={Journal of Computer Technology and Software},
  volume={4},
  number={4},
  year={2025}
}

@article{yang2024aligning,
  title={Aligning llms through multi-perspective user preference ranking-based feedback for programming question answering},
  author={Yang, Hongyu and He, Liyang and Hou, Min and Shen, Shuanghong and Li, Rui and Hou, Jiahui and Ma, Jianhui and Zhao, Junda},
  journal={arXiv preprint arXiv:2406.00037},
  year={2024}
}

@article{chakraborty2024maxmin,
  title={MaxMin-RLHF: Alignment with diverse human preferences},
  author={Chakraborty, Souradip and Qiu, Jiahao and Yuan, Hui and Koppel, Alec and Huang, Furong and Manocha, Dinesh and Bedi, Amrit Singh and Wang, Mengdi},
  journal={arXiv preprint arXiv:2402.08925},
  year={2024}
}

@article{das2024active,
  title={Active preference optimization for sample efficient rlhf},
  author={Das, Nirjhar and Chakraborty, Souradip and Pacchiano, Aldo and Chowdhury, Sayak Ray},
  journal={arXiv preprint arXiv:2402.10500},
  year={2024}
}

@article{wang2023lora,
  title={LoRA ensembles for large language model fine-tuning},
  author={Wang, Xi and Aitchison, Laurence and Rudolph, Maja},
  journal={arXiv preprint arXiv:2310.00035},
  year={2023}
}

@inproceedings{barreto2018transfer,
  title={Transfer in deep reinforcement learning using successor features and generalised policy improvement},
  author={Barreto, Andre and Borsa, Diana and Quan, John and Schaul, Tom and Silver, David and Hessel, Matteo and Mankowitz, Daniel and Zidek, Augustin and Munos, Remi},
  booktitle={International Conference on Machine Learning},
  pages={501--510},
  year={2018},
  organization={PMLR}
}

@article{barreto2017successor,
  title={Successor features for transfer in reinforcement learning},
  author={Barreto, Andr{\'e} and Dabney, Will and Munos, R{\'e}mi and Hunt, Jonathan J and Schaul, Tom and van Hasselt, Hado P and Silver, David},
  journal={Advances in neural information processing systems},
  volume={30},
  year={2017}
}

@article{gershman2012successor,
  title={The successor representation and temporal context},
  author={Gershman, Samuel J and Moore, Christopher D and Todd, Michael T and Norman, Kenneth A and Sederberg, Per B},
  journal={Neural Computation},
  volume={24},
  number={6},
  pages={1553--1568},
  year={2012},
  publisher={MIT Press One Rogers Street, Cambridge, MA 02142-1209, USA journals-info~…}
}

@article{fujimoto2023sale,
  title={For sale: State-action representation learning for deep reinforcement learning},
  author={Fujimoto, Scott and Chang, Wei-Di and Smith, Edward and Gu, Shixiang Shane and Precup, Doina and Meger, David},
  journal={Advances in neural information processing systems},
  volume={36},
  pages={61573--61624},
  year={2023}
}

@article{watter2015embed,
  title={Embed to control: A locally linear latent dynamics model for control from raw images},
  author={Watter, Manuel and Springenberg, Jost and Boedecker, Joschka and Riedmiller, Martin},
  journal={Advances in neural information processing systems},
  volume={28},
  year={2015}
}

@article{hafner2019dream,
  title={Dream to control: Learning behaviors by latent imagination},
  author={Hafner, Danijar and Lillicrap, Timothy and Ba, Jimmy and Norouzi, Mohammad},
  journal={arXiv preprint arXiv:1912.01603},
  year={2019}
}

@article{huang2024deal,
  title={Deal: Decoding-time alignment for large language models},
  author={Huang, James Y and Sengupta, Sailik and Bonadiman, Daniele and Lai, Yi-an and Gupta, Arshit and Pappas, Nikolaos and Mansour, Saab and Kirchhoff, Katrin and Roth, Dan},
  journal={arXiv preprint arXiv:2402.06147},
  year={2024}
}

@article{khanov2024args,
  title={Args: Alignment as reward-guided search},
  author={Khanov, Maxim and Burapacheep, Jirayu and Li, Yixuan},
  journal={arXiv preprint arXiv:2402.01694},
  year={2024}
}

@article{mudgal2023controlled,
  title={Controlled decoding from language models},
  author={Mudgal, Sidharth and Lee, Jong and Ganapathy, Harish and Li, YaGuang and Wang, Tao and Huang, Yanping and Chen, Zhifeng and Cheng, Heng-Tze and Collins, Michael and Strohman, Trevor and others},
  journal={arXiv preprint arXiv:2310.17022},
  year={2023}
}

@article{schulman2017proximal,
  title={Proximal policy optimization algorithms},
  author={Schulman, John and Wolski, Filip and Dhariwal, Prafulla and Radford, Alec and Klimov, Oleg},
  journal={arXiv preprint arXiv:1707.06347},
  year={2017}
}

@article{bakker2022fine,
  title={Fine-tuning language models to find agreement among humans with diverse preferences},
  author={Bakker, Michiel and Chadwick, Martin and Sheahan, Hannah and Tessler, Michael and Campbell-Gillingham, Lucy and Balaguer, Jan and McAleese, Nat and Glaese, Amelia and Aslanides, John and Botvinick, Matt and others},
  journal={Advances in neural information processing systems},
  volume={35},
  pages={38176--38189},
  year={2022}
}

@article{christiano2017deep,
  title={Deep reinforcement learning from human preferences},
  author={Christiano, Paul F and Leike, Jan and Brown, Tom and Martic, Miljan and Legg, Shane and Amodei, Dario},
  journal={Advances in neural information processing systems},
  volume={30},
  year={2017}
}

@article{garg2023extreme,
  title={Extreme q-learning: Maxent rl without entropy},
  author={Garg, Divyansh and Hejna, Joey and Geist, Matthieu and Ermon, Stefano},
  journal={arXiv preprint arXiv:2301.02328},
  year={2023}
}

@article{rafailov2024r,
  title={From $ r $ to $Q^*$: Your language model is secretly a $Q$-function},
  author={Rafailov, Rafael and Hejna, Joey and Park, Ryan and Finn, Chelsea},
  journal={arXiv preprint arXiv:2404.12358},
  year={2024}
}

@article{cui2025process,
  title={Process reinforcement through implicit rewards},
  author={Cui, Ganqu and Yuan, Lifan and Wang, Zefan and Wang, Hanbin and Li, Wendi and He, Bingxiang and Fan, Yuchen and Yu, Tianyu and Xu, Qixin and Chen, Weize and others},
  journal={arXiv preprint arXiv:2502.01456},
  year={2025}
}

@article{hazan2016introduction,
  title={Introduction to online convex optimization},
  author={Hazan, Elad and others},
  journal={Foundations and Trends{\textregistered} in Optimization},
  volume={2},
  number={3-4},
  pages={157--325},
  year={2016},
  publisher={Now Publishers, Inc.}
}

@inproceedings{yun-etal-2025-sleepless,
    title = "Sleepless Nights, Sugary Days: Creating Synthetic Users with Health Conditions for Realistic Coaching Agent Interactions",
    author = "Yun, Taedong  and
      Yang, Eric  and
      Safdari, Mustafa  and
      Lee, Jong Ha  and
      Kumar, Vaishnavi Vinod  and
      Mahdavi, S. Sara  and
      Amar, Jonathan  and
      Peyton, Derek  and
      Aharony, Reut  and
      PhD, Andreas Michaelides  and
      Schneider, Logan Douglas  and
      Galatzer-Levy, Isaac  and
      Jia, Yugang  and
      Canny, John  and
      Gretton, Arthur  and
      Mataric, Maja",
    editor = "Che, Wanxiang  and
      Nabende, Joyce  and
      Shutova, Ekaterina  and
      Pilehvar, Mohammad Taher",
    booktitle = "Findings of the Association for Computational Linguistics: ACL 2025",
    month = jul,
    year = "2025",
    address = "Vienna, Austria",
    publisher = "Association for Computational Linguistics",
    url = "https://aclanthology.org/2025.findings-acl.729/",
    doi = "10.18653/v1/2025.findings-acl.729",
    pages = "14159--14181",
    ISBN = "979-8-89176-256-5"
}

@ARTICLE{Yfantidou2022-ay,
  title     = "{LifeSnaps}, a 4-month multi-modal dataset capturing unobtrusive
               snapshots of our lives in the wild",
  author    = "Yfantidou, Sofia and Karagianni, Christina and Efstathiou,
               Stefanos and Vakali, Athena and Palotti, Joao and Giakatos,
               Dimitrios Panteleimon and Marchioro, Thomas and Kazlouski, Andrei
               and Ferrari, Elena and Girdzijauskas, {\v{S}}ar\={u}nas",
  journal   = "Sci. Data",
  publisher = "Springer Science and Business Media LLC",
  volume    =  9,
  number    =  1,
  pages     =  663,
  month     =  oct,
  year      =  2022,
  language  = "en"
}

@article{goldberg1992development,
  title={The development of markers for the Big-Five factor structure.},
  author={Goldberg, Lewis R},
  journal={Psychological assessment},
  volume={4},
  number={1},
  pages={26},
  year={1992},
  publisher={American Psychological Association}
}

\appendix
\newpage
\section{Derivations of Results in Section \ref{sec:universal_rep}}
\subsection{Proof of Lemma \ref{lem:linearly_parameterized_q}}
Given the reference LLM representation $\psi(s) \in \mathbb{R}^d$ and an action token embedding $\nu_{\text{ref}}(a) \in \mathbb{R}^d$, the reference policy is expressed as a softmax function:
$
\pi(a|s) = \exp(\psi(s)^\top \nu_{\text{ref}}(a))/\int_{b \in A} \exp(\psi(s)^\top \nu_{\text{ref}}(b)) db
$. We can also rewrite the inner product using the identity $\psi(s)^\top \nu_{\text{ref}}(a) = -\frac{1}{2}(\|\psi(s) - \nu_{\text{ref}}(a)\|^2 - \|\psi(s)\|^2 - \|\nu_{\text{ref}}(a)\|^2)$. Substituting this into the reference policy equation reveals a Gaussian kernel:
$
\pi_{\text{ref}}(a|s) \propto \exp\left(-\frac{1}{2}\|\psi(s) - \nu_{\text{ref}}(a)\|^2\right)
$, which measures the similarity between the state and action embeddings.
To create a more tractable linear representation, we approximate this Gaussian kernel using Random Fourier Features (RFF). RFF approximates a continuous shift-invariant kernel with an inner product of randomized feature maps. Applying this technique yields a spectral representation of the reference LLM policy:
\begin{equation}\label{eq:ebm_ref}
\pi_{\text{ref}}(a|s) = \frac{\langle \phi_\omega(s), \mu_{\text{ref},\omega}(a) \rangle_{N(\omega)}}{\langle \phi_\omega(s), \int_{b \in A} \mu_{\text{ref},\omega}(b) db \rangle_{N(\omega)}}
\end{equation}
Here, the expectation $\langle \cdot, \cdot \rangle_{N(\omega)}$ is over a random frequency vector $\omega \sim \mathcal{N}(0,I)$, and the feature maps are defined as:
$\phi_\omega(s) = \exp(-i\omega^\top \psi(s))\exp(\frac{1}{2}\|\psi(s)\|^2)$, $\mu_{\text{ref},\omega}(a) = \exp(i\omega^\top\nu_{\text{ref}}(a))\exp(\frac{1}{2}\|\nu_{\text{ref}}(a)\|^2)
$
This spectral representation transforms the non-linear kernel into a linear inner products in a randomized feature space, providing a direct path toward understanding the conditions under which the corresponding optimal Q-function can also be represented linearly.

To prove this technical lemma, we start by understanding the L-step decodability condition in Assumption \ref{assumption:l_step_decodability}.
Consider the optimal Bellman equation for $Q^*(s_{t}, a_{t})$ in Equation \ref{eq:soft_q}, which can be easily derived by unrolling the one-step Bellman equation forward in time for $L$ steps and using the time consistency property of exponential risk measure \cite{hau2023entropic} $\log\mathbb{E}_{\pi_{\text{ref}}}\exp({Q(s,a)}/{\beta})$, and the fact that the transition dynamics of the language MDP is deterministic $s'=(s,a)$:
\begin{equation}\label{eq:Q_star_exp}
Q^*(s_{t}, a_{t}) = r(s_t,a_t)+\beta\log\mathbb{E}_{\pi_{\text{ref},t+1:t+L-1}}\left[\exp\left(\frac{1}{\beta}\sum_{i=t+1}^{t+L-1} r(s_i, a_i) + Q^*(s_{t+L}, a_{t+L})\right)\right],\,\,\forall s_t,a_t.
\end{equation}
According to the $L$-step decodability assumption, the function $Q^*(s_{t+L}, a_{t+L})$ from the above formulation is directly independent to the earlier history $s_t=(s_0,a_0, a_1,\ldots, a_{t-1})$ and $a_t$.
However, the forward steps are conducted according to the policy $\pi_{\text{ref},t+1:t+L-1}=\{\pi_{\text{ref}}(\cdot|s_{t+1}),\ldots,\pi_{\text{ref}}(\cdot|s_{t+L-1})\}$ still depends on parts of $s_t$, hence the distribution in the above expectation still retains a dependence on $s_t$. 

To further study the dependence on history of this optimal Q-function, we introduce a final observation: For the reference policy $\pi_{\text{ref}}$ in Equation \eqref{eq:ebm_ref}, under the L-step decodability assumption there exists a corresponding policy $\kappa_{\pi_{\text{ref}}}$, known as the \emph{moment matching policy}, that conditions on a sufficient latent variable (the reference LLM representation) to generate the same expected observation dynamics while being independent of history older than $L$ steps \cite{zhang2023provable}. Now, with this observation we consider the $L$-step auto-regressive distribution induced by the reference policy $\pi_{\text{ref}}$, i.e., $\mathbb P^{\pi_{\text{ref}}}(s_{t+1:t+L}, a_{t+1:t+L}|s_t, a_t)$. Under $L$-step decodability, for any arbitrary state-action pair $(s_t,a_t)$ at any step $t$, this forward distribution emits the following spectral decomposition:
\begin{equation}\label{eq:l_forward_ref_ebm}
\mathbb P^{\pi_{\text{ref}}}(s_{t+1:t+L}, a_{t+1:t+L}|s_t, a_t) = \frac{\langle \phi_\omega(s_t, a_t), \mu_{\kappa_{\text{ref}},\omega}(s_{t+1:t+L}, a_{t+1:t+L}) \rangle_{\mathcal{N}(\omega)}}{\langle \phi_\omega(s_t, a_t), \int\mu_{\kappa_{\text{ref}},\omega}(s_{t+1:t+L}, a_{t+1:t+L})ds_{t+1:t+L}da_{t+1:t+L}\rangle_{\mathcal{N}(\omega)}},
\end{equation}
When modeling the state-action pairs in the next
$h\in\{1,\ldots,L\}$ steps, the above expression uses the \emph{moment matching} trick, which uses the policy $\kappa_{\text{ref}}$ that only depends on the latent variable and is independent of $(x_t, a_t)$. Under this policy, $\mu_{\kappa_{\text{ref}},\omega}(s_{t+1:t+L}, a_{t+1:t+L})$, is then a function that maps to the latent space by construction and is independent of history, leading to the above spectral representation. 
Utilizing this result, one can re-write Equation \eqref{eq:Q_star_exp} as
\[
\begin{split}
    &Q^*(s_{t}, a_{t})-r(s_t,a_t) \\
    =&\beta\log\int\mathbb P^{\pi_{\text{ref}}}(s_{t+1:t+L}, a_{t+1:t+L}|s_t, a_t)\exp\left(\frac{1}{\beta}\sum_{i=t+1}^{t+L-1} r(s_i, a_i) + Q^*(s_{t+L}, a_{t+L})\right)ds_{t+1:t+L}da_{t+1:t+L}\\
    =&\beta\log\frac{\int\langle \phi_\omega(s_t, a_t), \mu_{\kappa_{\text{ref}},\omega}(s_{t+1:t+L}, a_{t+1:t+L}) \rangle_{\mathcal{N}(\omega)}\exp\left(\frac{1}{\beta}\sum_{i=t+1}^{t+L-1} r(s_i, a_i) + Q^*(s_{t+L}, a_{t+L})\right)ds_{t+1:t+L}da_{t+1:t+L}}{\langle \phi_\omega(s_t, a_t), \int\mu_{\kappa_{\text{ref}},\omega}(s_{t+1:t+L}, a_{t+1:t+L})ds_{t+1:t+L}da_{t+1:t+L}\rangle_{\mathcal{N}(\omega)}}\\
=&\beta\log\frac{\int\langle \phi_\omega(s_t, a_t), \mu_{\kappa_{\text{ref}},\omega}(s_{t+1:t+L}, a_{t+1:t+L}) \exp(\frac{1}{\beta}\sum_{i=t+1}^{t+L-1} r(s_i, a_i) + Q^*(s_{t+L}, a_{t+L})) \rangle_{\mathcal{N}(\omega)}ds_{t+1:t+L}da_{t+1:t+L}}{\langle \phi_\omega(s_t, a_t), \int\mu_{\kappa_{\text{ref}},\omega}(s_{t+1:t+L}, a_{t+1:t+L})ds_{t+1:t+L}da_{t+1:t+L}\rangle_{\mathcal{N}(\omega)}}\\
=&\beta\log\frac{\bigg\langle \phi_\omega(s_t, a_t), \underbrace{\int\mu_{\kappa_{\text{ref}},\omega}(s_{t+1:t+L}, a_{t+1:t+L}) \exp\left(\frac{1}{\beta}\sum_{i=t+1}^{t+L-1} r(s_i, a_i) + Q^*(s_{t+L}, a_{t+L})\right)ds_{t+1:t+L}da_{t+1:t+L}}_{\mu_{\beta,Q^*,\omega}=\exp(-i\omega^\top \nu_{\beta,\sum r}) \exp(\|\nu_{\beta,\sum r}\|^2/2), \text{ for some $\nu_{\beta,\sum r}$ }} \bigg\rangle_{\mathcal{N}(\omega)}}{\langle \phi_\omega(s_t, a_t), \underbrace{\int\mu_{\kappa_{\text{ref}},\omega}(s_{t+1:t+L}, a_{t+1:t+L})ds_{t+1:t+L}da_{t+1:t+L}}_{\mu_{\overline{\kappa}_{\text{ref}},\omega}=\exp(-i\omega^\top \nu_{\text{ref}}) \exp(\|\nu_{\text{ref}}\|^2/2), \text{ for some $\nu_{\text{ref}}$ }}\rangle_{\mathcal{N}(\omega)}},
\end{split}
\]
where the first equality follows directly from the definitions of the optimal Q-function and the forward distribution in Equation \eqref{eq:l_forward_ref_ebm}, the second equality follows from the linear property of the Fourier spectral representation, which is characterized by an inner product $\langle \cdot, \cdot \rangle_{N(\omega)}$ in the frequency space, and the third equality follows from algebraic operations of the corresponding random Fourier feature maps. The above arguments further imply that 
\[
Q^*(s_{t}, a_{t})-r(s_t,a_t)=\beta\cdot\log \frac{\exp(\psi((s,a))^\top \nu_{\beta,\sum r})}{\exp(\psi((s,a))^\top \nu_{\text{ref}})} = \psi((s_t,a_t))^\top(\nu_{\beta,\sum r} - \nu_{\text{ref}})\cdot \beta,\,\,\forall s_t,a_t.
\]
This further implies that $Q^*(s, a)=\psi((s,a))^\top\nu_{\beta,r,\text{ref}}$
with $\nu_{\beta,r,\text{ref}}=\nu_{\text{r}} + \nu_{\beta,\sum r} - \nu_{\text{ref}}$, meaning that the optimal Q-function can be linearly parametrized with the reference LLM feature $\psi$ under these conditions, completing the proof of this lemma.

\newpage
\subsection{Proof of Theorem \ref{thm:q_soups}}
For any base and spectral soup temperatures $\beta,\beta'>0$, logit-mixture vector $\lambda\in\mathbb R^K$, recall the spectral soup policy $\tilde{\pi}_{\lambda}(a|s)\propto \pi_{\text{ref}}(a|s)\exp(\sum_k\lambda_k Q^*_k(s,a)/\beta')$, the personalized policy $\pi^*_{\mathbf{w}}(a|s)\propto \pi_{\text{ref}}(a|s)\exp(Q^*_{\mathbf{w}}(s,a)/\beta)$, i.e., $D_{\text{KL}}(\pi^*_{\mathbf{w}}(\cdot|s)||\tilde{\pi}_{\lambda}(\cdot|s))$, the personalized value function
\[
V^*_{\mathbf{w}}(s)=\max_{\pi} \mathbb{E}_{\pi} \left[ \sum_{t=0}^{T-1} r_{\mathbf{w}}(s_t, a_t) - \beta D_{KL} (\pi(\cdot | s_t) \| \pi_{\text{ref}}(\cdot | s_t)) |
s_0=s\right]=\beta\cdot\log\mathbb E_{a\sim \pi_{\text{ref}}(\cdot|s)} \left[\exp\frac{ Q^*_{\mathbf w}(s,a)}{\beta}\right],
\]
and notice that the spectral soup policy has the following optimal value function:
\[
V^{\lambda^*}_{\mathbf{w},\beta'}(s):=\max_{\lambda\in\mathbb R^K} \,\,\mathbb{E}_{\tilde\pi_\lambda} \left[ \sum_{t=0}^{T-1} r_{\mathbf{w}}(s_t, a_t) - \frac{\beta'}{\sum_k|\lambda_k|} \cdot D_{KL} (\tilde\pi_\lambda(\cdot | s_t) \| \pi_{\text{ref}}(\cdot | s_t)) |
s_0=s\right]\,\,\text{s.t.}\,\, \beta\sum_k|\lambda_k|\leq \beta'.
\]

First, we aim to compute an upper bound for the KL divergence between the spectral soup policy and the personalized policy, i.e., $D_{\text{KL}}(\pi^*_{\mathbf{w}}(\cdot|s)||\tilde{\pi}_{\lambda}(\cdot|s))$. Expanding the KL divergence term, we obtain the following inequality for any logit-mixture vector $\lambda\in\mathbb R^K$:
\begin{align}
&0\leq  D_{\text{KL}}(\pi^*_{\mathbf{w}}(\cdot|s)||\tilde{\pi}_{\lambda}(\cdot|s))=\mathbb E_{a\sim\pi^*_{\mathbf{w}}(\cdot|s)}\left[\log\frac{\pi^*_{\mathbf{w}}(a|s)}{\tilde{\pi}_{\lambda}(a|s)}\right]\nonumber\\
=&\mathbb E_{a\sim\pi^*_{\mathbf{w}}(\cdot|s)}\left[\log \pi_{\text{ref}}(a|s) + \frac{Q^*_{\mathbf{w}}(s, a)-V^*_{\mathbf{w}}(s)}{\beta'}-\left(\log \pi_{\text{ref}}(a|s)+\frac{\sum_k\lambda_k Q^*_k(s,a)}{\beta'}-\log\mathbb E_{a\sim \pi_{\text{ref}}(\cdot|s)} \exp\frac{\sum_k\lambda_k Q^*_k(s,a)}{\beta'}\right)\right]\nonumber\\
= &\mathbb E_{a\sim\pi^*_{\mathbf{w}}}\left[\frac{Q^*_{\mathbf{w}}(s, a)}{\beta}-\sum_k\frac{\lambda_k Q^*_k(s,a)}{\beta'}\right]+\left[\log\mathbb E_{a\sim \pi_{\text{ref}}(\cdot|s)} \exp\frac{\sum_k\lambda_k Q^*_k(s,a)}{\beta'}-\log\mathbb E_{a\sim \pi_{\text{ref}}(\cdot|s)} \exp\frac{ Q^*_{\mathbf w}(s,a)}{\beta}\right]\nonumber\\
\leq &\frac{1}{\beta'}\bigg(\|E_{a\sim\pi^*_{\mathbf{w}}}[\psi((s,a))]\|_2+\|E_{a\sim\pi_{\text{ref}}}|\psi((s,a))|\|_2\bigg)\cdot \left\|\frac{\beta'}{\beta}\nu_{\beta,r_{\mathbf w},\text{ref}}-\sum_k\lambda_k\, \nu_{\beta,r_k,\text{ref}}\right\|_2,
\end{align}
where the last inequality follows from Lemma \ref{lem:linearly_parameterized_q} and the Lipchitz property of the $\log\mathbb{E}_{\pi_{\text{ref}}}\exp(X)$ function (with Lipschitz constant $1$).
Second, we aim to derive the performance sub-optimality bound with respect to the personalized-soup value function $V^{\lambda^*}_{\mathbf{w},\beta'}(s)$, and personalized value function $V^*_{\mathbf{w}}(s)$. By the above definitions, we can easily argue that (i) $V^{*}_{\mathbf{w}}(s)\geq V^{\lambda^*}_{\mathbf{w},\beta'}(s)$ as the personalized-soup optimization problem always has a lower objective value than that of the personalized RLHF problem, and (ii)
$
V^{\lambda^*}_{\mathbf{w},\beta'}(s) \geq V^{\pi^*_k}_{\mathbf{w},\beta'}(s)$, $\forall k\in\{1,\ldots,K\}$, 
as the right hand side of this inequality can be recovered by setting the logit mixture vector $\lambda$ as the corresponding one-hot vector with magnitude $\beta'/\beta>0$ at the $k$-th attribute. Furthermore, based on the assumption of non-negative rewards $\mathbf r$, which implies that all optimal value functions ($V^*_{\mathbf w}$, $V^{\lambda^*}_{\mathbf w}$, $V^*_k$, but not necessarily $V^{\pi^*_k}_{\mathbf{w},\beta'}$ since it is not an optimal value function) are also non-negative. With ${\beta'}/{(\beta\sum_k|\lambda^*_k|)} \geq 1$, we can therefore express the sub-optimality performance bound of the personalized-soup policy at any state $s$ as
\begin{equation}
\begin{split}
 V^*_{\mathbf{w}}(s)\geq V^{\lambda^*}_{\mathbf{w},\beta'}&(s)\geq \sum_k\frac{|\lambda^*_k|}{\sum_k|\lambda^*_k|}V^{\pi^*_k}_{\mathbf{w},\beta'}(s)=\sum_k\frac{|\lambda^*_k|}{\sum_k|\lambda^*_k|}\left(V^{\pi^*_k}_{\mathbf{w},\beta'}(s)-V_k^*(s)+V_k^*(s)\right)\\
\geq&\underbrace{\sum_k\frac{|\lambda^*_k |}{\sum_k|\lambda^*_k|}\left(V^{\pi^*_k}_{\mathbf{w},\beta'}(s)-V_k^*(s)\right)}_{A(s)}+\frac{\beta}{\beta'}\Bigg(\underbrace{\sum_k \lambda^*_k V^*_k(s)-\beta\sum_k|\lambda^*_k|\log\mathbb E_{a\sim \pi_{\text{ref}}(\cdot|s)} \exp\frac{\sum_k\lambda^*_k Q^*_k(s,a)}{\beta\sum_k|\lambda^*_k|}}_{B(s)}\\
&\quad+\underbrace{\beta\sum_k|\lambda^*_k|\log\mathbb E_{a\sim \pi_{\text{ref}}(\cdot|s)} \exp\frac{\sum_k\lambda^*_k Q^*_k(s,a)}{\beta\sum_k|\lambda^*_k|}-\beta' \log\mathbb E_{a\sim \pi_{\text{ref}}(\cdot|s)} \exp\frac{ Q^*_{\mathbf w}(s,a)}{\beta}}_{C(s)}\Bigg)+V^*_{\mathbf{w}}(s).\nonumber
\end{split}
\end{equation}
This performance bound can further be simplified via the following sequence of inequalities:
\begin{align}
&  V^{\lambda^*}_{\mathbf{w},\beta'}(s)\geq V^*_{\mathbf{w}}(s)+A(s)+({\beta}/{\beta'})B(s) - 
\|E_{a\sim\pi_{\text{ref}}}|\psi((s,a))|\|_2\|\nu_{\beta,r_{\mathbf w},\text{ref}}-({\beta}/{\beta'})\sum_k\lambda^*_k \nu_{\beta,r_k,\text{ref}}\|_2\nonumber\\
\geq & V^*_{\mathbf{w}}(s)-\sum_{t=0}^{T-1}\mathbb E_{\underline{\pi}}\left[\|\psi((s_t,a_t))\|_2|s\right]\left\|\nu_{r_{\mathbf{w}}}-\frac{\sum_k|\lambda^*_k|\nu_{r_k}}{\sum_k|\lambda^*_k|}\right\|_2+\frac{\beta}{\beta'}B(s) - \|E_{a\sim\pi_{\text{ref}}}|\psi((s,a))|\|_2\|\nu_{\beta,r_{\mathbf w},\text{ref}}-\frac{\beta}{\beta'}\sum_k\lambda^*_k \nu_{\beta,r_k,\text{ref}}\|_2\nonumber\\
\geq & V^*_{\mathbf{w}}(s)-\sum_{t=0}^{T-1}\mathbb E_{\underline{\pi}}\left[\|\psi((s_t,a_t))\|_2|s_0=s\right]\cdot\left\|\nu_{r_{\mathbf{w}}}-\frac{\sum_k|\lambda^*_k|\nu_{r_k}}{\sum_k|\lambda^*_k|}\right\|_2+\frac{\beta^2}{\beta'}\cdot\sum_k\Delta_k(s)(\lambda^*_k)_-\nonumber\\
&\quad\quad\quad\quad\quad\quad\quad- 
\|E_{a\sim\pi_{\text{ref}}}|\psi((s,a))|\|_2 \cdot\|\nu_{\beta,r_{\mathbf w},\text{ref}}-({\beta}/{\beta'})\sum_k\lambda^*_k\, \nu_{\beta,r_k,\text{ref}}\|_2,
\label{eq:v_suboptimal}
\end{align}
where the first inequality follows from the derivations in Equation \eqref{eq:kl_bdd}, i.e., Lipschitz continuity of $\log\mathbb{E}_{\pi_{\text{ref}}}\exp(X)$, and the convexity of  the function $f(x)=x^{\beta'/(\beta\sum_k|\lambda^*_k|)}$, $\beta'/(\beta\sum_k|\lambda^*_k|)\geq 1$ i.e.,
\[
\begin{split}
C(s) =&\beta\sum_k|\lambda^*_k|\log\mathbb E_{a\sim \pi_{\text{ref}}(\cdot|s)} \exp\frac{\sum_k\lambda^*_k Q^*_k(s,a)}{\beta\sum_k|\lambda^*_k|}-\beta' \log\mathbb E_{a\sim \pi_{\text{ref}}(\cdot|s)} \exp\frac{ Q^*_{\mathbf w}(s,a)}{\beta}\\
=&\beta\sum_k|\lambda^*_k|\log\mathbb E_{a\sim \pi_{\text{ref}}(\cdot|s)} \left(\exp\frac{\sum_k\lambda^*_k Q^*_k(s,a)}{\beta'}\right)^{\beta'/\beta\sum_k|\lambda^*_k|}-\beta' \log\mathbb E_{a\sim \pi_{\text{ref}}(\cdot|s)} \exp\frac{ Q^*_{\mathbf w}(s,a)}{\beta}\\
\geq&\frac{\beta'\cdot\beta\sum_k|\lambda^*_k|}{\beta\sum_k|\lambda^*_k|}\log\left(\mathbb E_{a\sim \pi_{\text{ref}}(\cdot|s)} \exp\frac{\sum_k\lambda^*_k Q^*_k(s,a)}{\beta'}\right)-\beta' \log\mathbb E_{a\sim \pi_{\text{ref}}(\cdot|s)} \exp\frac{ Q^*_{\mathbf w}(s,a)}{\beta}\\
\geq & - \|E_{a\sim\pi_{\text{ref}}}|\psi((s,a))|\|_2\|\cdot\|\frac{\beta'}{\beta}\nu_{\beta,r_{\mathbf w},\text{ref}}-\sum_k\lambda^*_k\, \nu_{\beta,r_k,\text{ref}}\|_2,
\end{split}
\]
$\forall s$, the second inequality follows from utilizing the definitions of value functions in the soft MDP:  
\[
\begin{split}
&A(s):=\sum_k\frac{|\lambda^*_k |}{\sum_k|\lambda^*_k|}(V^{\pi^*_k}_{\mathbf{w},\beta'}(s)-V_k^*(s))=\frac{1}{\sum_k|\lambda^*_k|}\sum_k|\lambda^*_k|\,\mathbb E_{\pi_{ k}^*}\left[\sum_{t=0}^{T-1}\psi((s_t,a_t))|s_0=s\right]^\top\left(\nu_{r_{\mathbf{w}}}-\nu_{r_k}\right)\\
\geq &\frac{1}{\sum_k|\lambda^*_k|}\min_{\pi}\mathbb E_{\pi}\left[\sum_{t=0}^{T-1}\psi((s_t,a_t))^\top\sum_k|\lambda^*_k|\left(\nu_{r_{\mathbf{w}}}-\nu_{r_k}\right)|s_0=s\right]\\
= & \frac{1}{\sum_k|\lambda^*_k|}\mathbb E_{\underline{\pi}}\left[\sum_{t=0}^{T-1}\psi((s_t,a_t))^\top\sum_k|\lambda^*_k|\left(\nu_{r_{\mathbf{w}}}-\nu_{r_k}\right)|s_0=s\right]
\geq -\sum_{t=0}^{T-1}\mathbb E_{\underline{\pi}}\left[\|\psi((s_t,a_t))\|_2|s\right]\cdot\left\|\nu_{r_{\mathbf{w}}}-\frac{\sum_k|\lambda^*_k|\nu_{r_k}}{\sum_k|\lambda^*_k|}\right\|_2,
\end{split}
\]
and assume both the linear parameterization rewards in Assumption \ref{assumption:linear_reward} and the existence of a minimal policy over the mixture of reward differences $\underline\pi\in\argmin_{\pi}\mathbb E_{\pi}\left[\sum_{t=0}^{T-1}\psi((s_t,a_t))^\top\sum_k|\lambda^*_k|\left(\nu_{r_{\mathbf{w}}}-\nu_{r_k}\right)|s_0=s\right]$,  
and the third inequality follows from a hypothesis that there exists $\Delta_k(s)\geq 0$ such that the following property holds (such a technical result will be derived in the remaining part of the proof):
\begin{equation}\label{eq:mixed_v}
B(s):=\sum_k \lambda^*_k V^*_k(s)-\beta\sum_k|\lambda^*_k|\log\mathbb E_{a\sim \pi_{\text{ref}}(\cdot|s)} \exp\frac{\sum_k\lambda^*_k Q^*_k(s,a)}{\beta\sum_k|\lambda^*_k|}\geq \beta\cdot\underbrace{\sum_k\Delta_k(s)(\lambda^*_k)_-}_{\leq 0}. 
\end{equation}

To show that Equation \eqref{eq:mixed_v} holds, first consider the following upper-bound of $1/x$, with $0\leq \underline x\leq x \leq \overline x$, with respect to an arbitrary anchor point $\underline x\leq a\leq \overline x$, such that $|x-a|\leq |a|$, $\forall x$:
\[
\begin{split}
&\frac{1}{x}=\frac{1}{a} - \frac{x-a}{a^2} + \frac{(a-x)^2}{a^2}\frac{1}{x}\leq \frac{1}{a} - \frac{x-a}{a^2} + \frac{(a-\underline{x})^2}{a^2}\frac{1}{x}\\
\implies& \frac{1}{x}\leq\sum_{k=0}^\infty (\frac{(a-\underline{x})^2}{a^2})^k\left(\frac{1}{a} - \frac{x-a}{a^2}\right)=\frac{2a-x}{a^2}\frac{1}{1-\frac{(a-\underline{x})^2}{a^2}}=\frac{2a-x}{a^2}\frac{a^2}{\underline x (2a-\underline x)}=\frac{2a-x}{\underline x (2a-\underline x)}.
\end{split}
\]
Substituting to the above inequality 
\[
\begin{split}
&x= \exp\frac{Q^*_k(s,a)}{\beta},\quad \underline x=\underline{M}(s) \cdot\mathbb E_{a\sim\pi_{\text{ref}}}\left[\exp\frac{Q^*_k(s,a)}{\beta}\right],\\
&\overline x=\overline{M}(s) \cdot\mathbb E_{a\sim\pi_{\text{ref}}}\left[\exp\frac{Q^*_k(s,a)}{\beta}\right], \quad a=M(s)\cdot\mathbb E_{a\sim\pi_{\text{ref}}}\left[\exp\frac{Q^*_k(s,a)}{\beta}\right],
\end{split}
\]
for the importance-sampling (IS) factor $M(s):=\pi^*_k(a|s)/\pi_{\text{ref}}(a|s)$, IS lower bound $\underline M(s)$, and IS upper bound $\overline M(s)$, where $0\leq\underline{M}(s)\leq 1$, $\overline{M}(s)\geq 1$, $M(s)\in[\underline{M}(s),\overline{M}(s)]$, and taking expectation $\mathbb E_{a\sim\pi_{\text{ref}}}$ on both sides, this expression becomes
\begin{align}
&\mathbb E_{a\sim \pi_{\text{ref}}(\cdot|s)}\left[\frac{1}{\exp\frac{Q^*_k(s,a)}{\beta}}\right]\leq \frac{2M(s)-1}{\underline{M}(s) (2M(s)-\underline{M}(s))}\frac{1}{\mathbb E_{a\sim \pi_{\text{ref}}(\cdot|s)} \left[\exp\frac{Q^*_k(s,a)}{\beta}\right]}\label{eq:neg_logsumexp}\\
\implies& \log \mathbb E_{a\sim \pi_{\text{ref}}(\cdot|s)}\left[\frac{1}{\exp\frac{Q^*_k(s,a)}{\beta}}\right] \leq \log \frac{2M(s)-1}{\underline{M}(s) (2M(s)-\underline{M}(s))} -\log\mathbb E_{a\sim \pi_{\text{ref}}(\cdot|s)} \left[\exp\frac{Q^*_k(s,a)}{\beta}\right] \nonumber\\
\implies& -\log \mathbb E_{a\sim \pi_{\text{ref}}(\cdot|s)}\left[\exp\frac{-Q^*_k(s,a)}{\beta}\right] \geq \log\mathbb E_{a\sim \pi_{\text{ref}}(\cdot|s)} \left[\exp\frac{Q^*_k(s,a)}{\beta}\right]-\underbrace{\log \frac{2M(s)-1}{\underline{M}(s) (2M(s)-\underline{M}(s))}}_{\Delta^M_k(s)}.\nonumber
\end{align}
The above bound is only valid when $\Delta^M_k(s)\geq 0$. We will achieve that and also minimize $\Delta^M_k(s)$ by choosing the right $M(s)$. Notice that, since $0\leq\underline{M}(s)\leq 1$ and $1\leq\overline{M}(s)$, the expression
$
\exp\Delta^M_k(s)=\frac{2M(s)-1}{\underline{M}(s) (2M(s)-\underline{M}(s))} \geq 1
$
only when $M(s) \geq \frac{1}{2} (1 + \underline{M}(s))$, and this value would monotonically increase with $M$ beyond that. However, to satisfy the necessary condition $-|a|\leq x-a\leq |a|$, $\forall x$, that guarantees the convergence of the above geometric sum, for $\underline x\leq x\leq \overline x$, the smallest $a$ value can only be $(\overline x+\underline x)/2$, or in other words the smallest possible $M$ is $(\overline{M}(s)+\underline{M}(s))/2$, which is a valid choice as it is greater than $(1+\underline{M}(s))/2$. Substituting $M^*=(\overline{M}(s)+\underline{M}(s))/2$ into $\Delta^M_k(s)$ yields 
\begin{equation}\label{eq:delta_k}
\Delta_k(s):=\Delta^{M^*}_k(s)=\log (\overline{M}(s)+\underline{M}(s) - 1) - \log(\overline{M}(s)\cdot\underline{M}(s))\geq 0.
\end{equation}
If $\Delta_k(s)$ is constructed via Equation \eqref{eq:delta_k}, then the inequality in Equation \eqref{eq:mixed_v} can be proven by using the convexity arguments and definitions of $V^*_k$ via the following inequalities:
\begin{align}
\frac{1}{\beta\sum_k|\lambda^*_k|}\sum_k \lambda^*_k V^*_k(s) 
= &\frac{1}{\beta\sum_k|\lambda^*_k|}\sum_k \lambda^*_k\beta \log\mathbb E_{a\sim \pi_{\text{ref}}(\cdot|s)} \left[\exp\frac{Q^*_k(s,a)}{\beta}\right]\nonumber\\
\geq &\frac{1}{\beta\sum_k|\lambda^*_k|}\sum_k \left(\beta\lambda^*_k(\text{sgn}(\lambda^*_k)) \log\mathbb E_{a\sim \pi_{\text{ref}}(\cdot|s)} \left[\exp\frac{\text{sgn}(\lambda^*_k)Q^*_k(s,a)}{\beta}\right] + \beta\Delta_k(s)(\lambda^*_k)_-\right)\nonumber\\
=&\frac{\beta}{\beta}\sum_k \left(\frac{|\lambda^*_k|}{\sum_k|\lambda^*_k|} \log\mathbb E_{a\sim \pi_{\text{ref}}(\cdot|s)} \left[\exp\frac{\text{sgn}(\lambda^*_k)Q^*_k(s,a)}{\beta}\right]+ \frac{\Delta_k(s)(\lambda^*_k)_-}{\sum_k|\lambda^*_k|}\right)\nonumber\\
\geq&\log\mathbb E_{a\sim \pi_{\text{ref}}(\cdot|s)} \left[\exp\frac{\sum_k|\lambda^*_k|\text{sgn}(\lambda^*_k) Q^*_k(s,a)}{\beta \sum_k|\lambda^*_k|}\right]+ \frac{\sum_k\Delta_k(s)(\lambda^*_k)_-}{ \sum_k|\lambda^*_k|}\nonumber\\
=&\log\mathbb E_{a\sim \pi_{\text{ref}}(\cdot|s)} \left[\exp\frac{\sum_k\lambda^*_k Q^*_k(s,a)}{\beta \sum_k|\lambda^*_k|}\right]+ \frac{\sum_k\Delta_k(s)(\lambda^*_k)_-}{ \sum_k|\lambda^*_k|},\nonumber
\end{align}
where the first equality follows from the definition of $V^*_k(s)$, the first inequality follows from the arguments in Equation \eqref{eq:neg_logsumexp} applied to the cases when $\lambda^*_k<0$ (with equality holds trivially when $\lambda^*_k\geq 0$), the second and third equalities follows from simple algebra, the second inequality follows from convex property of the function $\log\mathbb{E}_{\pi_{\text{ref}}}\exp(X)$.

Combining all these arguments complete the proof of this theorem.

\newpage
\section{Experiment Details} \label{sec:exp_details}
\subsection{Experimental Domains}
To assess the effectiveness of our approach, we conduct
empirical evaluations of spectral souping on three realistic LLM personalization experiments. Each experiment contain two phases, an offline learning phase of specialized policies and an online adaptation phase for LLM personalization.

The first experiment is built upon the \textbf{UltraFeedback dataset}, which initially contains a training set of 60,829 examples and a test set of 985 examples. Each data point consists of an input prompt $x$ and a pair of distinct agent responses, $(y_1, y_2)$, where each response $y_i$ is annotated with a four-dimensional feature vector, $\phi(x, y_i)$, quantifying its helpfulness, honesty, instruction-following, and truthfulness. We leverage this fine-grained scores to synthesize diverse preference labels, each with a unique weight vector $\mathbf{w} \in \mathbb{R}^4$. Pairwise preference rankings of two responses is determined by the dot product of their feature vectors with this weight vector; for instance, $y_1$ is preferred over $y_2$ if $\langle \phi(x, y_1), \mathbf{w} \rangle > \langle \phi(x, y_2), \mathbf{w} \rangle$. In the offline phase, we build a library of $K=30$ specialized policies. We generate 30 distinct datasets by creating 30 unique preference vectors, $\mathbf{w}_k$, each sampled from a distribution centered around a basis vector for one of the four attributes (e.g., $[1,0,0,0]$ for helpfulness). The online phase evaluates the algorithm's generalization to novel preferences under ambiguous conditions. On the same dataset of prompts and responses, we simulate responses of eight held-out "users," proxied by publicly available reward models. Crucially, the preference functions of these models were unseen during offline training, providing a rigorous test of generalization. To further amplify the task's difficulty, we filter the dataset to retain only the most contentious examples where preferences conflict, resulting in a benchmark of $23,614$ training and $401$ test examples, forcing the model to learn nuanced preference trade-offs. 

The second experimental setup focuses on \textbf{personalized text-to-image (T2I)} generation within the PASTA framework, which involves a 5-turn (H=5) interactive process. At each turn, the agent presents the user with a 4x4 slate of 16 images, where each column corresponds to a unique prompt expansion. The core generation models include Stable Diffusion XL for images and Gemini 1.5 Flash for creating a candidate set of 25 prompts, from which the four are selected for the slate. The utility functions are based on fine-tuned Gemma 2B models. For the offline learning phase, we generated K=32 specialized datasets from over 30,000 simulated user rollouts, totaling more than 2.5 million images. These rollouts were guided by 32 distinct user models designed to capture myopic, turn-by-turn preferences. In this simulation, a user provides an absolute satisfaction score (on a 5-point scale) for the best image in each column based on its relevance to the original prompt. The choice is then modeled as selecting the column that received the highest score. In the online phase, the algorithm's adaptability is tested against 5 held-out, simulated users. Each of these "auto-raters" is powered by a unique, pre-trained Q-function that models a holistic, session-level preference. In contrast to the myopic offline models, these Q-functions evaluate the entire 5-turn session based on salient user values, such as aesthetic quality, prompt-image consistency, or a specific artistic style. These user-mimicking Q-functions were trained on a large offline dataset using Implicit Q-Learning.

Our third experiment is grounded in the healthcare domain of \textbf{sleep coaching}. We begin by obtaining detailed user profiles from 68 real individuals from the LifeSnaps dataset. Each profile is constructed using a rich set of attributes, including demographics (age, gender), health metrics (BMI, average and variable sleep duration). From these attributes, we generate a 'sleep profile' for each user, detailing their primary sleep concern, goals, and barriers. To simulate conversations, each user is instantiated as an LLM whose prompt contains the user's entire backstory vignette, along with the full preceding conversation history, ensuring that the dialogue is consistently grounded in the user's profile. For the offline learning phase, we generate $K=15$ specialized preference datasets, each corresponding to one of the five personality dimensions. To create each dataset, we first generate $1,000$ pairs of 10-turn conversations by having our synthetic users interact with the coaching agent—a 'Talker' and 'Reasoner' system powered by Gemini 1.5 Pro. The key step for creating specific data is via persona-based ranking: each of the $1,000$ conversation pairs is ranked by a reward function that specifically embodies one of the five target personalities. For the online adaptation phase, we evaluate the algorithm's performance on a test set of $512$ samples for each of 5 distinct "users," who are simulated by a auto-rater system with generative feedback. This system employs Gemini 1.5 Flash to score conversations by systematically evaluating them against a detailed set of rubrics designed to holistically assess sleep-coaching quality. These rubrics cover the agent's tone and style (friendliness, supportiveness, empowerment), its ability to understand the user (rapport establishment, capturing concerns, efficient information gathering), and the efficacy of its personalized intervention (collaborative goal-setting, relevance, and quality of the structured plan).

\subsection{Model and Training Details}
\subsection{Additional Results}\label{appendix:results}

\begin{figure}[htbp]
    \centering
    \includegraphics[width=\textwidth]{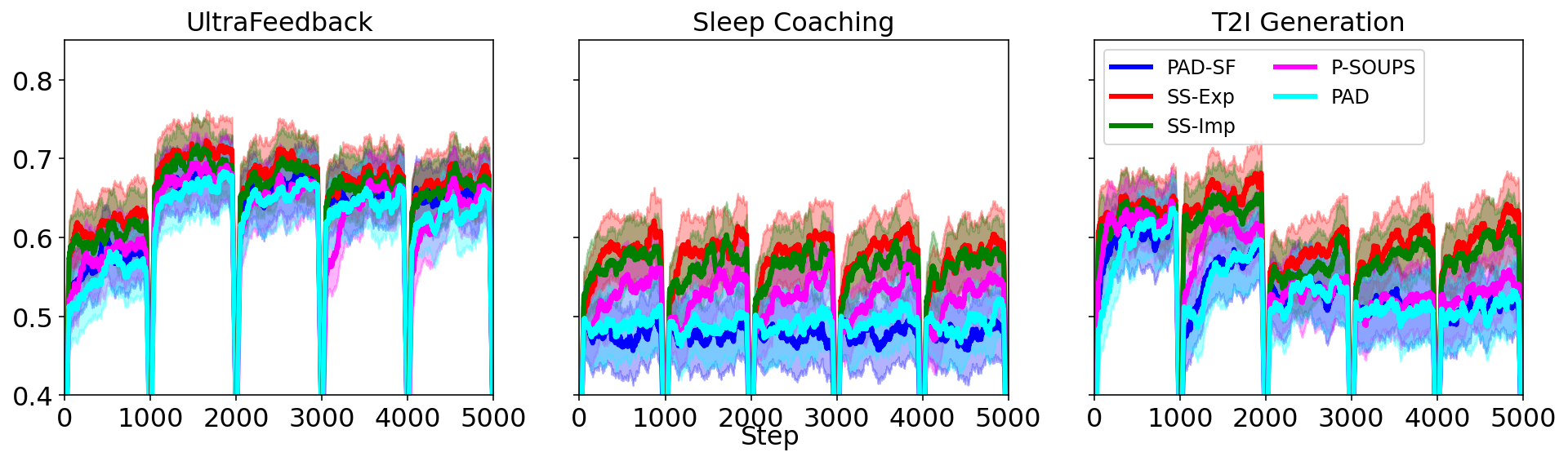} 
    \caption{Test-time Training Performance of Different Methods using the Gemma3 1B model: Explicit \& Implicit Spectral Souping (SS-Exp \& SS-Imp), P-SOUPS, PAD, PAD-SF, RLHF, adapted to 5 various users in the UltraFeedback, T2I Generation, Sleep Coaching domains. The SS methods (especially SS-Exp) consistently and outperform P-SOUPS and the PAD baselines, demonstrating superior performance in online adaptation.}
    \label{fig:test_time_training_1B}
\end{figure}
\begin{figure}[htbp]
    \centering
    \includegraphics[width=\textwidth]{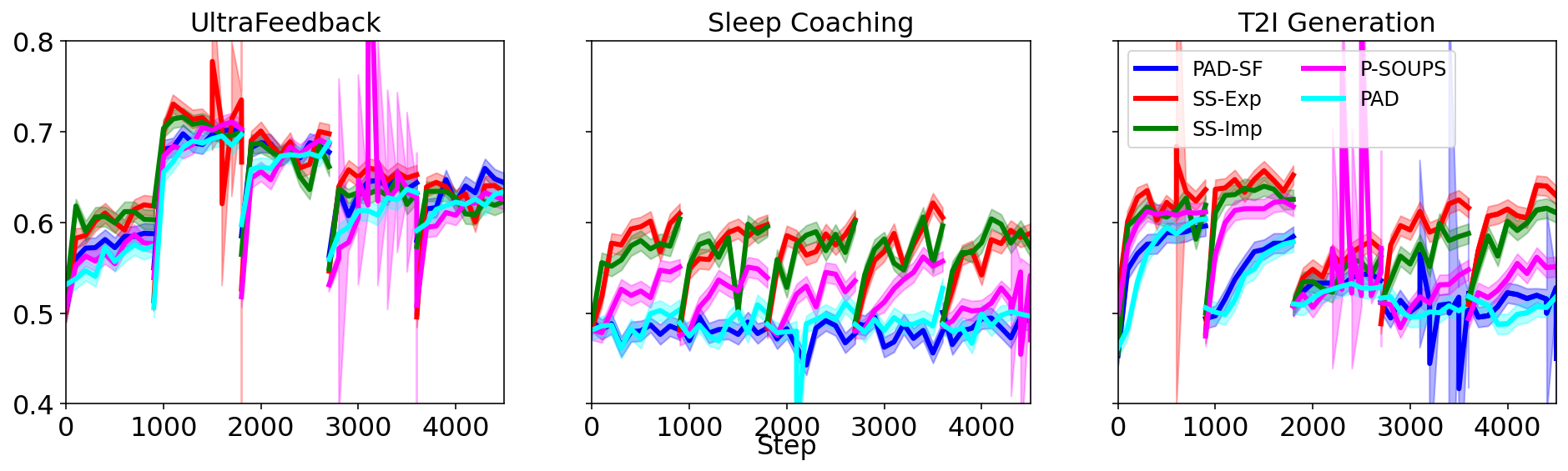} 
    \caption{Evaluation Performance of Different Online Adaptation Methods using the Gemma3 1B model: Explicit \& Implicit Spectral Souping (SS-Exp \& SS-Imp), P-SOUPS, PAD, PAD-SF, RLHF, across adapted to 5 various users in the UltraFeedback, T2I Generation, Sleep Coaching domains. The superior performance of the SS methods (over P-SOUPS, PAD, and PAD-SF baselines) is also generalizable to online evaluations.}
    \label{fig:test_time_eval_1B}
\end{figure}

\newpage
\section{Sequential Estimation of $\lambda$}

The loss functions in both cases (binary labels and preference labels) are convex in $\lambda$, and so inference of these parameters is tractable using off-the-shelf convex optimization methods. However evaluation of the loss function is relatively expensive and must be performed for each user, ideally in real-time. Thus, in this section we propose an online method optimization of $\lambda$. The formulation of our method is based on sequential variational inference, but we will show the resulting inference algorithm is a simple weighted least squares update. 

Our variational method aims to compute a variational posterior $q(\lambda)$ via 
\begin{equation}
    q^*(\lambda) = \argmin \E_{q(\lambda)}[\mathcal{L}(\lambda; \mathcal{B})] + \KL(q(\lambda) \mid p(\lambda)).
\end{equation}
We fix $q(\lambda) = \N(\bar{\lambda}, S)$, and a Gaussian prior $p(\lambda)$. Note that this objective is convex in the parameters of the variational posterior. In the case where we sequentially observe trajectories and labels at interaction $n$, $(\tau_n, l_n)$, we propose a sequential method for identification of $\lambda$ based on variational continual learning. In particular, at time index $n$, we compute variational posterior $q_n$ via
\begin{equation}
    q_n(\lambda) = \argmin_{q} \E_{q(\lambda)}[\mathcal{L}(\lambda; (\tau_n, l_n))] + \KL(q(\lambda) \mid q_{n-1}(\lambda)) \label{eq:svi_obj}.
\end{equation}
Briefly, this method aims to sequentially infer the variational posterior by regularizing to the posterior of the previous timestep, using a method similar to sequential filtering. 

We derive the least squares update for the preference learning case, and the binary classification model is a straightforward extension. We approximate \eqref{eq:svi_obj} via second order Taylor expansion, allowing exact computation of the expectation. We will define $\Delta_{n-1} = \frac{R_{n-1}(w) - R_{n-1}(l)}{\beta}$ (where each $R$ term here is vectorized over $k$). This term is computed by the log likelihoods of the policy as in Section \ref{sec:offline_learning_specialized}. We further define $\sigma_{n-1} = \sigma(\bar{\lambda}_{n-1}^\top \Delta_{n-1})$, the predictive preference likelihood after updating at step $n-1$.

Computing the analytical minimum of this second order approximation, we get updates
\begin{align}
    S_n^{-1} &= S_{n-1}^{-1} + \sigma_{n-1} (1-\sigma_{n-1}) \Delta_{n-1} \Delta_{n-1}^\top\\
    \bar{\lambda}_{n} &= \bar{\lambda}_{n-1} + (1-\sigma_{n-1}) S \Delta_{n-1}
\end{align}
which are inexpensive to compute compared to the cost of the model evaluation. The updates may be made cheaper by exploiting rank-1 updates, but this is a relatively minor consideration. This update can naively be performed once per timestep/user feedback, corresponding to a Newton-style step per interaction. It can also be performed multiple times (matching iteratively reweighted least squares) by setting $\sigma_{n-1} = \sigma(\bar{\lambda}_{n}^\top \Delta_{n-1})$.






\end{document}